\documentclass[10pt,twocolumn,letterpaper]{article}

\usepackage{iccv}              %

\definecolor{iccvblue}{rgb}{0.21,0.49,0.74}
\usepackage[pagebackref,breaklinks,colorlinks,allcolors=iccvblue]{hyperref}
\usepackage{algorithm}
\usepackage{algpseudocode}
\usepackage{listings}
\usepackage{xcolor}
\usepackage{booktabs}
\usepackage{multirow}
\usepackage{makecell}
\usepackage{placeins}
\usepackage{colortbl}

\definecolor{TeacherBlack}{RGB}{128, 128, 128}
\definecolor{PrunedYellow}{RGB}{255, 255, 67}
\definecolor{RewardBlue}{RGB}{106, 171, 221}
\definecolor{SFTGreen}{RGB}{104, 192, 111}
\definecolor{DPOred}{RGB}{225, 140, 126}
\definecolor{ReDPOpurple}{RGB}{189, 137, 189}

\def\paperID{15522} %
\def\confName{ICCV}
\def\confYear{2025}

\title{\includegraphics[height=1.0em]{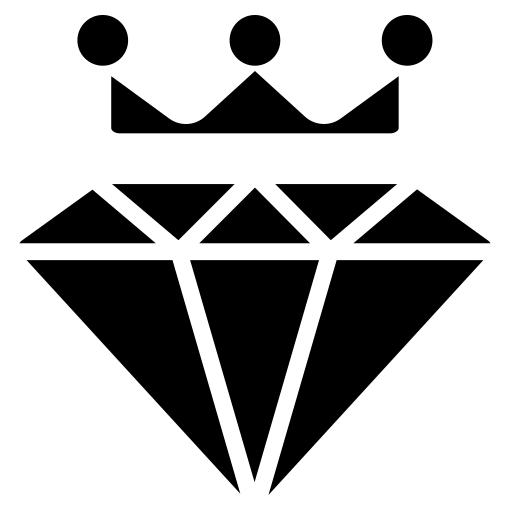} V.I.P. : Iterative Online Preference Distillation for \\ Efficient Video Diffusion Models}

\author{Jisoo Kim \hspace{3mm} Wooseok Seo \hspace{3mm} Junwan Kim \hspace{3mm} Seungho Park \hspace{3mm} Sooyeon Park \hspace{3mm} Youngjae Yu\\
Yonsei University\\
\texttt{\small \{jisoo6687, justin\_seo, jw1510, gomi0904, dreamyou070, yjy\}@yonsei.ac.kr}
}

\begin{document}

\twocolumn[{
    \maketitle
    \centering
    \bigskip
}]

\begin{abstract}
With growing interest in deploying text-to-video (T2V) models in resource-constrained environments, reducing their high computational cost has become crucial, leading to extensive research on pruning and knowledge distillation methods while maintaining performance. However, existing distillation methods primarily rely on supervised fine-tuning (SFT), which often leads to mode collapse as pruned models with reduced capacity fail to directly match the teacher’s outputs, ultimately resulting in degraded quality. To address this challenge, we propose an effective distillation method, \loss, that integrates DPO and SFT. Our approach leverages DPO to guide the student model to focus on recovering only the targeted properties, rather than passively imitating the teacher, while also utilizing SFT to enhance overall performance. We additionally propose \ours, a novel framework for filtering and curating high-quality pair datasets, along with a step-by-step online approach for calibrated training. We validate our method on two leading T2V models, VideoCrafter2 and AnimateDiff, achieving parameter reduction of 36.2\% and 67.5\% each, while maintaining or even \emph{\textit{surpassing}} the performance of full models. Further experiments demonstrate the effectiveness of both \loss and \ours framework in enabling efficient and high-quality video generation. Our code and videos are available at https://jiiiisoo.github.io/VIP.github.io/.
\end{abstract}

\section{Introduction}
\label{sec:intro}

Recent advancements in video generation models have significantly improved their ability to produce high-fidelity and temporally coherent videos. However, these models typically require substantial computational costs and large memory footprints, which can be prohibitive for resource-constrained deployment. In particular, deploying on mobile phones or edge devices with strict memory and speed constraints often becomes infeasible with models of this scale.

\label{fig:teaser}
\begin{figure}[t] \centering
    \includegraphics[width=0.48\textwidth]{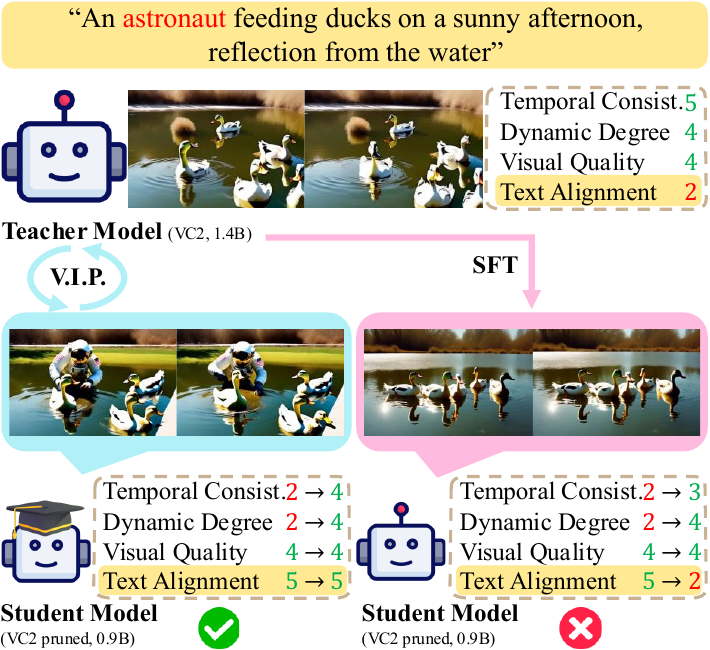}
    \caption{\textbf{Conceptual visualization of generated videos and benchmark scores from pruned models distilled using \ours and SFT}. Only the model trained with \ours generates correct concept (\textit{astronaut}) with high-quality video, indicating strong text alignment, even surpassing the full model. \textit{This indicates that \ours selectively improves \textcolor[rgb]{0.9,0,0}{red} (weak) dimensions while preserving \textcolor[rgb]{0,0.7,0}{green} (strong) ones, whereas SFT blindly mimics teacher, degrading even the previously better-performing aspects (text alignment).}}
    \label{fig:motivation}
\end{figure}

To address these challenges, recent research has explored strategies such as pruning~\cite{he2017channel, lagunas2021block, liu2017learning, xia2022structured} to develop lightweight models. However, since pruned models typically suffer from performance degradation compared to full models, knowledge distillation~\cite{hinton2015distilling} has emerged as a prominent approach for preserving strong generative performance, where a smaller student model learns to approximate the output of a larger teacher model, achieving comparable performance while reducing computational cost.

Conventional knowledge distillation methods for diffusion models rely largely on na\"ively imitating the teacher's outputs without addressing the limited capacity of the student model~\cite{kim2024bk, castells2024edgefusion, wu2024snapgen}. These approaches transfer knowledge in a direct but unstructured manner, forcing the student model to replicate the teacher’s behavior indiscriminately. Due to the inherent capacity gap, such direct distillation—often performed by minimizing the distance between the teacher’s and student’s predictions or features—tends to result in a suboptimal results~\cite{mse_bad_2_pmrf, mse_bad_3_wasserstein, mse_bad_perception_distortion}, as the student model lacks the expressiveness to fully reproduce the teacher’s outputs~\cite{lin2024sdxl_lightning}. Therefore, refining the learning objective is essential to ensure that the student allocates its limited capacity toward capturing critical generative patterns rather than blindly mimicking the teacher’s outputs.

As a novel alternative to conventional methods with direct supervision, we propose leveraging preference learning, commonly represented by Direct Preference Optimization (DPO)~\cite{rafailov2023direct}, to improve diffusion model distillation. Unlike conventional approaches that enforce strict alignment with the teacher’s outputs across all aspects, preference learning allows the student to selectively improve properties that degrade due to pruning while avoiding the unnecessary expenditure of its limited capacity on those that undergo only minor deterioration or even improve. Therefore, we formulate diffusion distillation as a preference learning task, where the teacher’s outputs are treated as winning responses and the student’s as losing responses. By learning from contrastive feedback, rather than direct supervision alone, the student model can prioritize recovering the most critical generative patterns that are negatively affected by pruning, rather than attempting to mimic the teacher indiscriminately. Given the unpredictability of pruning effects—where certain generative properties degrade while others persist or even strengthen~\cite{su2024f3, wu2024individual}—this approach enables a more adaptive and efficient transfer of knowledge.

To fully capitalize on the benefits of preference learning in diffusion model distillation, we introduce \ours, a step-by-step online distillation framework designed to iteratively guide the student model through progressive pruning. Unlike conventional one-shot pruning, which abruptly removes multiple model components and forces the student to compensate for drastic capacity loss, our approach adopts an iterative pruning strategy. At each stage, we selectively remove specific blocks, allowing the model to gradually adapt while maintaining generative capability. This step-by-step adaptation, by progressively updating the student model at each iteration rather than all at once, ensures that the student model is consistently updated and enhanced, enabling it to continually generate improved training data at each iteration. Consequently, this iterative approach not only mitigates the significant degradation typically resulted by one-shot pruning but also contributes to synthesizing high-quality preference pairs via improved student models.

Overall, our contributions are as follows:
\begin{itemize}
    \item We present a novel distillation loss \textbf{\loss} that integrates Direct Preference Optimization (DPO) into the diffusion framework, making it the \textit{first} to employ preference learning for pruned diffusion models.

    \item We propose \textbf{\ours}, a framework that incorporates an on-policy data curation strategy and an online distillation method, allowing pruned models to recover lost features more effectively. By structuring pruning and distillation in stages, our approach ensures stable optimization and improved generative performance.

    \item Through extensive experiments, we demonstrate that our method significantly outperforms existing distillation approaches, even surpassing the full model in performance, while reducing parameters by 36.2\% in VideoCrafter and 67.5\% in the AnimateDiff motion module.
\end{itemize}

\section{Related Works}
\subsection{Text to Video Generation} 
Text-to-video generation has gained considerable attention due to its broad potential for various applications. Early methods relied on traditional generative models such as Generative Adversarial Networks (GANs)~\cite{goodfellow2020generative} and Variational Autoencoders (VAEs)~\cite{kingma2013auto}, but these approaches often produced low-quality, short videos. With the success of image diffusion models~\cite{ho2020denoising, podell2023sdxl, rombach2022ldm}, recent methods extend them to video diffusion models by introducing additional temporal layers on top of the spatial layers in existing 2D diffusion models~\cite{blattmann2023stablevideodiffusion, blattmann2023videoldm, chen2024videocrafter2, guo2023animatediff, li2024t2v}. This approach effectively carries over the strong generative capabilities of image diffusion models to the video domain. However, while these advancements have enabled high-resolution video generation, their substantial computational costs, limiting practical use for real-world applications. To address this, we propose a distillation-based approach that reduces model size while keeping generative capabilities largely intact. Unlike prior methods that focus only on naive pruning or supervised fine-tuning, we introduce a targeted and adaptive mechanism to prioritize and recover the most critical generative properties sacrificed by aggressive pruning.

\subsection{Diffusion Distillation} 
Diffusion models are computationally intensive, requiring pruning-based optimization for efficiency. While pruning removes redundancy with minimal performance loss, it often requires further training. Some works~\cite{wu2024snapgen, li2023snapfusion, yahia2024mobile, zhao2024mobilediffusion, kim2024bk} finetune pruned models with diffusion loss, which requires substantial data and computation. BK-SDM~\cite{kim2024bk} introduces knowledge distillation~\cite{hinton2015distilling} by minimizing the distance between noise predictions of pruned and original models.  However, as such exact matching is limited due to reduced capacity~\cite{lin2024sdxl}, it further incorporates feature-level guidance. Recent work~\cite{wu2024individual} improves upon this by incorporating adversarial loss, which sharpens distributions to mitigate capacity limitations. Yet, adversarial methods inherently lack precise control over where sharpness is applied, often leading to unintended distortions~\cite{lin2024sdxl_lightning, lin2024animatediff_lightning} and mode collapse~\cite{xiao2021tackling}, posing stability challenges in practical applications. Moreover, only few methods explore such techniques for text-to-video diffusion, wherein temporal consistency and motion fidelity become challenge and prone to degeneration under pruning. To address this, we propose a stepwise pruning with a novel distillation process that (i) explicitly identifies the properties that the pruned model struggles with and (ii) guides the student to prioritize them in a more adaptive, preference-driven manner.

\subsection{Preference Alignment Training}
\label{sec:DPOrelatedwork}

Preference learning is widely used to align generative models, especially large language models (LLMs), with human preferences~\cite{ziegler2019fine,ouyang2022training}. Traditional methods train a separate reward model using human preference data, which subsequently guides model refinement via reinforcement learning (RL) algorithms such as Proximal Policy Optimization (PPO)~\cite{schulman2017proximal}. Recently, Direct Preference Optimization (DPO)~\cite{rafailov2023direct} emerged as a more streamlined alternative, bypassing explicit reward model training and directly optimizing models against human preferences on pairwise datasets. The simplicity of its training process has popularized DPO, leading to various adaptations across text~\cite{azar2024general,ethayarajh2024kto,hong2024orpo,meng2025simpo}, images~\cite{wallace2024diffusion,yang2024using,ren2025refining}, and videos~\cite{liu2024videodpo,zhang2024onlinevpo,jiang2025huvidpo}. Despite its widespread use, the application of preference learning in diffusion distillation remains largely understudied, due to the complexity of aligning iterative denoising steps with human preferences while maintaining generative capability. To the best of our knowledge, we are the \emph{first} to tackle these challenges by introducing a preference-guided framework tailored for diffusion distillation. In doing so, we seamlessly integrate preference alignment with iterative pruning and online distillation, enabling our method to continuously curate training data at each stage and effectively remedy newly emerging weaknesses in the pruned model.

\section{Method}
In this section, we propose \textbf{\ours} (\textbf{V}ideo diffusion distillation via \textbf{I}terative \textbf{P}reference learning), our distillation framework designed to efficiently transfer generative capabilities from high-capacity teacher diffusion models to their lightweight students. Motivated by limitations of standard distillation methods, we first highlight the need for explicit guidance into loss function (Sec.~\ref{subsec:motivation}). We then describe two key building blocks, pruning algorithm (Sec.~\ref{subsec:pruning}) and data curation (Sec.~\ref{subsec:data_curation}). Finally, we present \ours (Sec.~\ref{subsec:vip}), which integrates our proposed loss \textbf{\loss}, and step-by-step distillation to better reallocate the student’s limited capacity toward essential generative properties.
\subsection{Motivation}
\label{subsec:motivation}
\begin{figure}[thbp] \centering
    \includegraphics[width=0.48\textwidth]{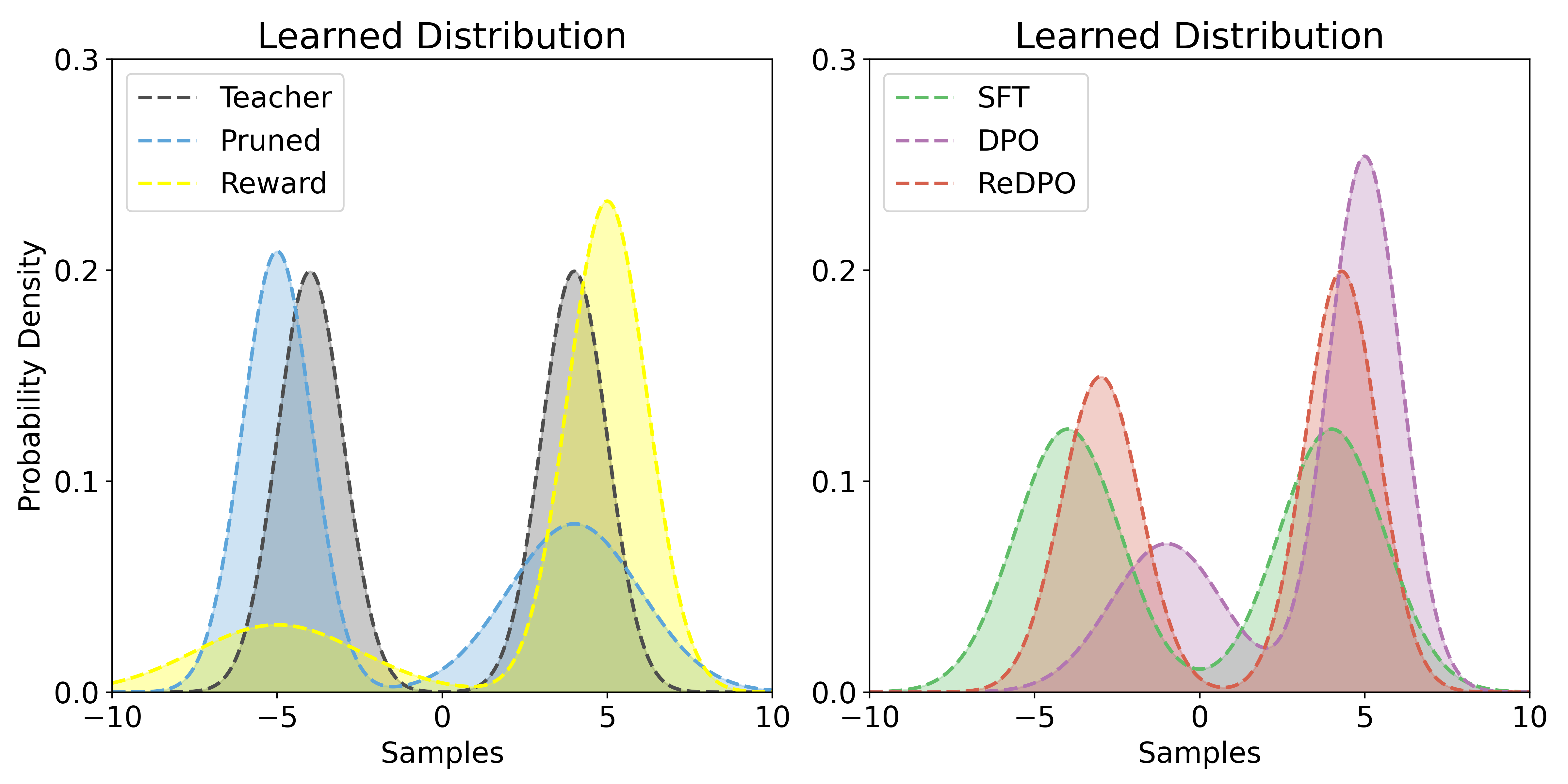}
    \caption{\textbf{Conceptual illustration of learned distributions from teacher and student models.} Conventional distillation methods (SFT) result in overly smooth distributions in low-capacity students. Our proposed method (ReDPO) effectively reallocates the student’s limited capacity toward the critical mode while preventing over-optimization.} \label{fig:motivation}
\end{figure}
Figure~\ref{fig:motivation} illustrates how a well-designed distillation loss helps capacity-constrained model effectively learn essential generative properties. Conventional distillation methods typically transfer knowledge by minimizing the $L_2$ distance between teacher and student predictions or feature representations~\cite{kim2024bk, castells2024edgefusion, wu2024snapgen}, known as supervised fine-tuning (SFT). While this approach guarantees convergence in sufficiently expressive models, it fails to do so when applied to models with limited capacity. In such cases, SFT loss often leads to a distributional averaging, where student produces samples that do not exist in teacher’s distribution. This occurs since minimizing SFT loss inherently prioritizes reducing overall error over preserving fine-grained details~\cite{mse_bad_perception_distortion, mse_bad_3_wasserstein}, resulting in an oversmoothed generative process (\textcolor{SFTGreen}{green} curve).

To address this, explicit guidance is required to help the student allocate its limited capacity to the most essential aspects of generation. We employ Direct Preference Optimization (DPO)~\cite{rafailov2023direct} to guide the student toward selectively meaningful generative properties, rather than mimiking all aspects of the teacher indiscriminately and wasting capacity. This prevents the capacity-limited student model from collapsing to a distributional average and enables effective use of constrained capacity. Notably, pruned student models often exhibit selective degradation (\textcolor{RewardBlue}{blue} curve) where some generative properties deteriorate while others remain unaffected or even improve. Using DPO, we can explicitly steer the student toward recovering these degraded properties rather than passively approximating the teacher’s distribution, ensuring a targeted and efficient distillation process.

\begin{figure*}[t] 
    \centering
    \includegraphics[width=\textwidth,keepaspectratio]{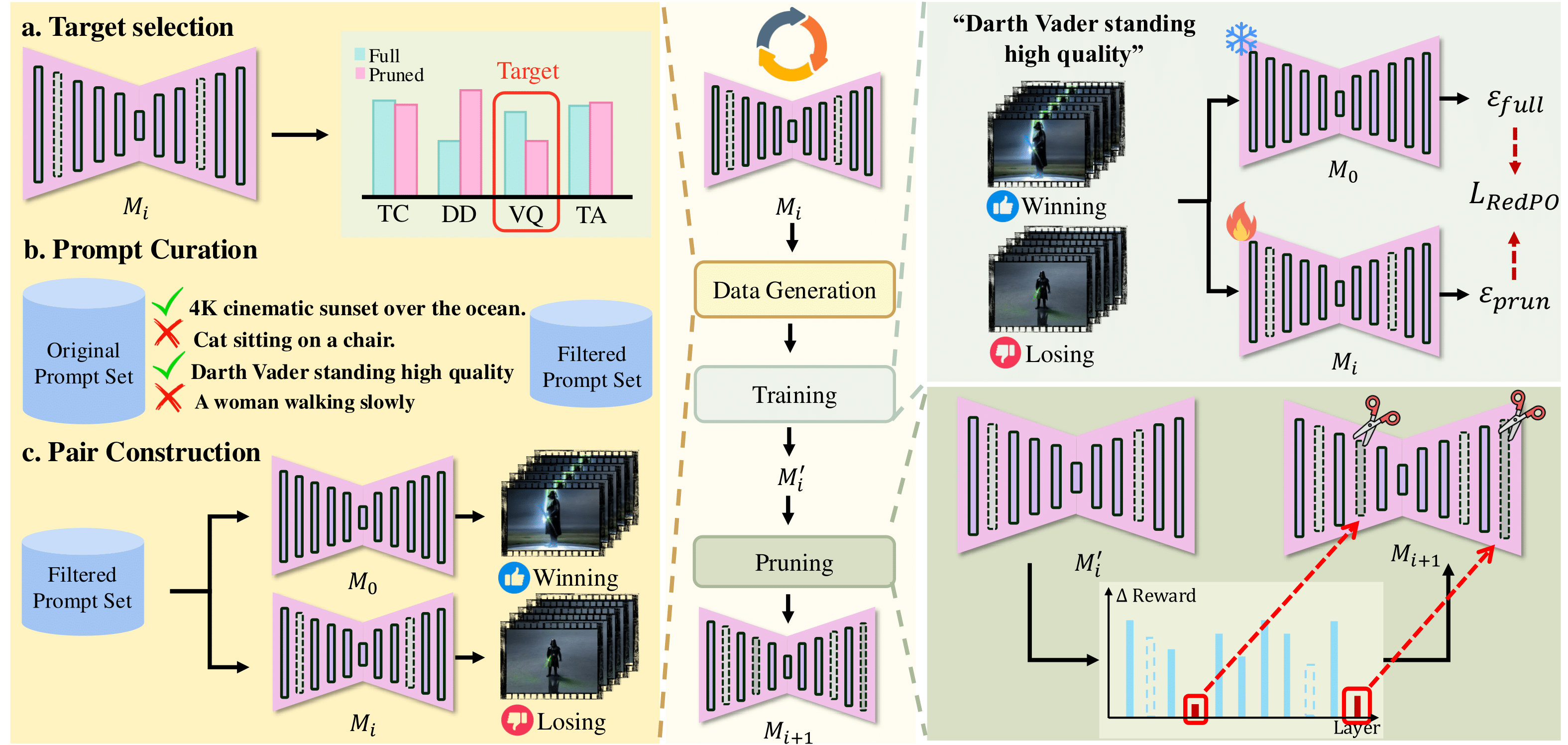}
    \caption{\textbf{Overall architecture}. Starting from a baseline model $M_{0}$, we obtain a pruned model $M_{1}$. Through a systemic evaluation \& preference data synthesis, and a training process using ReDPO, we obtain a preference-learned model $M_{1}'$. Then, $M_{1}'$ is pruned again to obtain $M_{2}$, which will run through an iterative distillation process \ours. Note that the teacher model is fixed to $M_{0}$, while the student model $M_{i}$ is dynamically updated.}
    \label{fig:figure2}
\end{figure*}

While DPO ideally enables the student model to allocate capacity more efficiently, a critical challenge is its inherent tendency to over-optimize~\cite{fisch2024robust, liu2024provably, pal2024smaug}. This arises from the objective, maximizing the \textit{relative} likelihood (i.e., the margin) between preferred and unpreferred responses. This can lead to an imperfect reward distribution, particularly in uncertain regions, resulting degraded output quality and bias toward out-of-distribution samples~\cite{liu2024provably} (\textcolor{ReDPOpurple}{purple} curve).

To address this, we introduce the SFT loss as regularizer, encouraging the student to mimic the teacher without relying solely on learned reward distributions. This combination results in a more balanced distillation process, avoiding inefficient resource allocation while preventing over-optimization. Building on this motivation, we propose \loss, which successfully transfers knowledge from the teacher to the student (\textcolor{DPOred}{red} curve). Section~\ref{subsubsec:loss} provides a detailed explanation of \loss, while Supplementary Section~\ref{sec:additional_motivation} presents toy experiment validating of our motivation.

\subsection{Pruning Algorithm}
\label{subsec:pruning}

For pruning blocks of model, we first evaluate the impact by removing each block individually using VideoScore's~\cite{he2024videoscore} total score and select blocks that have minimal impact compared to the full model. Here, we identify the properties that show the performance drop relative to the full model as the \textbf{target properties} for recovery. After each training step, as shown in Algorithm \ref{alg:progressive_pruning} in the Supplementary Section~\ref{sec:sup_pruning_algorithm}, the same pruning process is repeated as part of our step-by-step approach. This allows the model to progressively adapt to the full model's distribution, starting from easier settings by structurally pruning less impactful modules first, thereby facilitating a more effective learning process.
\label{subsec:dpo_limitation}
\subsection{Data Curation}
\label{subsec:data_curation}
We perform dataset filtering in two phases: \textit{prompt filtering} and \textit{video filtering}. 
Given the strong impact of prompt quality and semantics on video generation~\cite{cheng2025vpo,gao2025devil,liao2024evaluation}, we first filter prompts for quality and relevance to the targeted attribute. Using the filtered set, we curate videos by generating and evaluating outputs from both full and pruned models, and form winning-losing pairs based on target property.

\paragraph{Prompt filtering.}
First, we select high-quality prompts suitable for video inference and well-aligned with the targeted property. Given long prompts are not well-suited for video generation, following \cite{he2024videoscore}, we retain prompts with 5 to 25 words and remove articles to maintain a balanced length. We then filter prompts that contain the targeted property through LLM-based filtering, ensuring they explicitly describe relevant aspects that rule-based methods often miss. For instance, to target dynamic degree, we select prompts including motion-related elements, such as object movement or camera motion. Further details on LLM prompting are provided in Supplementary Section~\ref{sec:prompt_filter}.
\paragraph{Video curation.}
Using the filtered prompts, we generate a set of videos with both the full model and the pruned model, then evaluate them with VideoScore, which serves as a reward model. Based on the resulting scores, we construct training pairs \( (v_{\text{full}}, v_{\text{pruned}}) \) where the teacher outperforms the student for the targeted property \textit{p}, and also impose a minimum threshold \( \tau_p \) to prevent the inclusion of excessively low-quality samples.:
\[
S(v_{\text{full}}) > S(v_{\text{pruned}})> \tau_p, \quad v_{\text{full}} \in V_{\text{full}}^p, \quad v_{\text{pruned}} \in V_{\text{pruned}}^p
\]
where \( S(v) \) represents the reward of video \( v \).
This ensures that the winning sample is of high quality while effectively capturing cases where the pruned model underperforms. Further details are provided in Supplementarty Section~\ref{sec:data_curation_details}

\subsection{Iterative Preference Distillation}
\label{subsec:vip}
Building on a pruned model and curated data, our proposed method, \ours, introduces two key components: \textit{integration of SFT into DPO} and \textit{step-by-step online DPO learning}.

As noted in Section~\ref{subsec:motivation}, despite the advantages, DPO is prone to overoptimization, which can paradoxically degrade the output probability of winning responses. To address this, we integrate SFT into DPO as regularizer. Moreover, standard DPO operates offline, relying entirely on static datasets. In contrast, online methods like PPO~\cite{schulman2017proximal} sample training data and update policy iteratively, and have been shown to outperform offline approaches~\cite{tajwar2024preference,tang2024understanding,dong2024rlhf,ivison2025unpacking}, motivating recent proposals for online DPO variants~\cite{guo2024direct,chen2024self}. Inspired by theses findings, and to avoid the degradation from one-shot pruning which leads to drastic capacity loss, we propose step-by-step online DPO learning that incrementally optimizes the student model using updated samples at each stage. To these ends, we propose \loss (\textbf{Re}gularized \textbf{D}iffusion \textbf{P}reference \textbf{O}ptimization) and \ours (\textbf{V}ideo diffusion distillation via \textbf{I}terative \textbf{P}reference Optimization), described in the following sections.

\subsubsection{\loss}
\label{subsubsec:loss}

We enhance diffusion DPO loss~\cite{wallace2024diffusion} by incorporating SFT on preferred pairs as a regularizer, motivated by \cite{liu2024provably}. While the KL term in DPO imposes some constraints, it is insufficient to fully prevent overoptimization. SFT explicitly aligns student with distribution of preferred samples, reinforcing preference probability more effectively and improving generation quality. 

We set $\pi_{ref}$ as the full model and $\pi_{\theta}$ as the pruned model. The key idea behind \loss is that, given curated dataset, it selectively align the pruned model $\pi_{\theta}$ with the full model $\pi_{ref}$, while preserving aspects where the pruned model may already outperform the teacher.

To achieve this, objective $L_{diff-dpo}(\theta)$ is as follows :
\begin{equation}
\begin{aligned}
    &L_{diff-dpo}(\theta) = \\
    &- \mathbb{E}_{(x_0^w, x_0^l) \sim \mathcal{D}, \, t \sim \mathcal{U}(0,T), \, x_t^w \sim q(x_t^w \mid x_0^w), \, x_t^l \sim q(x_t^l \mid x_0^l)} \\
    &\quad \log \sigma ( -\beta T \omega(\lambda_t) ( \\
    &\quad \quad \| \boldsymbol{\epsilon}^w - \boldsymbol{\epsilon}_\theta(\boldsymbol{x}_t^w, t) \|_2^2 - \| \boldsymbol{\epsilon}^w - \boldsymbol{\epsilon}_{\text{ref}}(\boldsymbol{x}_t^w, t) \|_2^2 \\
    &\quad \quad - ( \| \boldsymbol{\epsilon}^l - \boldsymbol{\epsilon}_\theta(\boldsymbol{x}_t^l, t) \|_2^2 - \| \boldsymbol{\epsilon}^l - \boldsymbol{\epsilon}_{\text{ref}}(\boldsymbol{x}_t^l, t) \|_2^2 ) ) )
\end{aligned}
\end{equation}
Here, $\epsilon_\theta(\boldsymbol{x}_t^w, t)$ and $\epsilon_{ref}(\boldsymbol{x}_t^w, t)$ denote the noise predicted by $\pi_{\theta}$ and $\pi_{ref}$, respectively. The $SFT$ regularization term on the preferred pair is defined as:
\begin{equation}
\begin{aligned}
    &L_{SFT}(\theta) = \| \boldsymbol{\epsilon}_\theta(\boldsymbol{x}_t^w, t) - \boldsymbol{\epsilon}_{\text{ref}}(\boldsymbol{x}_t^w, t) \|_2^2
\end{aligned}
\end{equation}
Therefore, our final \loss loss is as follows :
\begin{equation}
\begin{aligned}
    &L_{\loss}(\theta) = L_{diff-dpo}(\theta) + w_{SFT}L_{SFT}(\theta)
\end{aligned}
\end{equation}
 $w_{SFT}$ is the weight of SFT loss. Furthermore, although \loss was specifically utilized for distillation task in this work, we emphasize that it can be applied robustly for general diffusion preference alignment purposes. We analyze the effect of $w_{SFT}$ in Supplementary Section~\ref{subsec:sftweight}.

\subsubsection{\ours}
\label{subsubsec:method}
The overall workflow of \ours is shown in Figure~\ref{fig:figure2}. We begin by pruning the full (teacher) model $M_{0}$ into a smaller model $M_{1}$(Section~\ref{subsec:pruning}). Both models ($M_{0}$ and $M_{1}$) generate videos for the same targeted prompts (Section~\ref{subsec:data_curation}), forming winning (from $M_{0}$) and losing (from $M_{1}$) pairs. These pairs are used to distill knowledge from $M_{0}$ into the pruned model via our preference-based distillation loss \loss, resulting in trained pruned model, $M_{1}'$. This refinement process repeats iteratively. The refined model $M_{1}'$ further pruned to $M_{2}$. In each subsequent iteration, the full model $M_{0}$ continues to produce the winning samples, while the current pruned model (e.g., $M_{2}$) generates updated losing samples based on targeted prompts identified through systematic evaluation of its deficiencies. These dynamically updated pairs form the training data for the next round of distillation. This iterative, online distillation cycle ensures pruned models to progressively adapt by consistently targeting and mitigating their latest performance weaknesses.

\section{Experiments}
In this section, we outline our evaluation approach. Section~\ref{subsec:experimental_settings} describes the experimental setup, covering benchmarks, baseline models, datasets, and hyperparameters. Section~\ref{subsec:experimental_results} presents both quantitative and qualitative analyses across various settings, comparing \ours, SFT-based distillation, and the full (teacher) model. An ablation study in Section~\ref{subsec:ablation} highlights the impact of each components. Further experimental details are in Supplementary Section~\ref{sec:sup_experimental_details}.

\begin{table*}[t]
    \centering
    \scalebox{0.89}{
    \begin{tabular}{llc|cccc|c|c|c}
        \toprule
        \textbf{Model} & \textbf{Stage} & \textbf{Method} & 
        \makecell{\textbf{Visual} \\ \textbf{Quality}} & 
        \makecell{\textbf{Temporal} \\ \textbf{Consistency}} & 
        \makecell{\textbf{Dynamic} \\ \textbf{Degree}} & 
        \makecell{\textbf{Text} \\ \textbf{Alignment}} & 
        \textbf{Average} & \textbf{Param(B)}\\
        \midrule
        \multirow{5}{*}{VideoCrafter 2} & & Full & 2.627 & 2.602 & \textbf{2.728} & 2.491 & 2.613 & 1.413 \\
        \cmidrule{2-9}
        & \multirow{2}{*}{Stage 1} & Pruned & \cellcolor{yellow!20} 2.609 & 2.588 & 2.744 & 2.487 & 2.609 & \multirow{2}*{1.174} \\
        &                        & \loss (ours) & \cellcolor{yellow!20} 2.630 & 2.608 & 2.731 & 2.510 & 2.620 & \\
        \cmidrule{2-9}
        & \multirow{2}{*}{Stage 2} & Pruned & 2.627 & \cellcolor{yellow!20} 2.595 & 2.725 & 2.486 & 2.608 & \multirow{2}*{0.902} \\
        &                        & \loss (ours) & \textbf{2.629} & \cellcolor{yellow!20}\textbf{2.617} & \textbf{2.728} & \textbf{2.518} & \textbf{2.623} \textcolor{blue}{(+0.010)} & \\
        \midrule
        \multirow{7}{*}{AnimateDiff} & & Full & \textbf{2.575} & 2.505 & 2.684 & 2.486 & 2.563 & 0.453 \\
        \cmidrule{2-9}
        & \multirow{2}{*}{Stage 1} & Pruned & \cellcolor{yellow!20} 2.561 & 2.494 & 2.713 & 2.488 & 2.564 & \multirow{2}*{0.309} \\
        &                        & \loss (ours) & \cellcolor{yellow!20} 2.579 & 2.524 & 2.685 & 2.499 & 2.572 & \\
        \cmidrule{2-9}
        & \multirow{2}{*}{Stage 2} & Pruned & 2.553 & \cellcolor{yellow!20} 2.478 & 2.718 & 2.470 & 2.555 & \multirow{2}*{0.219} \\
        &                        & \loss (ours) & 2.583 & \cellcolor{yellow!20} 2.525 & 2.688 & 2.496 & 2.573 & \\
        \cmidrule{2-9}
        & \multirow{2}{*}{Stage 3} & Pruned & 2.552 & \cellcolor{yellow!20} 2.469 & 2.736 & 2.505 & 2.566 & \multirow{2}*{0.147} \\
        &                        & \loss (ours) & 2.569 & \cellcolor{yellow!20} \textbf{2.513} & \textbf{2.695} & \textbf{2.496} & \textbf{2.568} \textcolor{blue}{(+0.005)} & \\
        \bottomrule
    \end{tabular}}
    \caption{\textbf{VideoScore results across stages with \ours applied to two baseline models}. The yellow-highlighted scores indicate our target training property, while the blue numbers represent the average improvement achieved by \ours compared to the full model. Our method not only successfully recovers performance lost due to pruning but also consistently surpasses the full model across nearly all evaluation criteria for both baselines. Our approach achieves these enhancements while reducing the parameter count by between 36.2\% and 67.5\%.}
    \label{tab:main_ours_comparison}
\end{table*}

\subsection{Settings}
\label{subsec:experimental_settings}
\paragraph{Evaluation details.}
We evaluated our models using VideoScore~\cite{he2024videoscore} and VBench~\cite{huang2024vbench}. VideoScore is human preference-aligned model trained on human-annotated dataset, while VBench is preference-aligned for aesthetics but uses rule-based for other attributes. We used the official test sets provided by each models. To improve motion evaluation, we propose a revised dynamic score for VBench. Noted in \cite{liao2025evaluation}, high dynamic scores are not always ideal-some prompts naturally require static motion, and generating appropriate stillness can reflect better model understanding. However, many static prompts are included in VBench's dynamic set, making it difficult to fairly assess motion quality. To address this, we manually re-annotated the dynamic test set, labeling prompts as either static or dynamic. For static prompts, we assigned a dynamic score of zero, ensuring that models are rewarded for correctly generating still motion when appropriate.
\paragraph{Training details.}
We use AnimateDiff~\cite{guo2023animatediff} and VideoCrafter~\cite{chen2024videocrafter2} as baselines. Since both are trained on WebVid-10M~\cite{bain2021frozen}, we filter prompts from this dataset to ensure distilled knowledge lies in pretrained training distribution. For pruning, since AnimateDiff uses a frozen Stable Diffusion 1.5~\cite{rombach2022high} U-Net as its backbone and only trains the motion module, we pruned the motion module exclusively. For VideoCrafter, we pruned entire blocks of U-Net. Both models were trained with $\beta$ = 5000 and learning rate = 6e-6, following the setting of VideoDPO~\cite{liu2024videodpo}. To match the scale of the DPO loss, we set the SFT weight to 1e4 for AnimateDiff and 1e6 for VideoCrafter. All experiments were conducted with a 2k prompt subset, batch size 2, and 2 training epochs per stage on 4 A100 GPUs.

\begin{table*}[h]
    \centering
    \scalebox{0.89}{
    \begin{tabular}{lc|cc|ccccccc}
        \toprule
        \textbf{Model} & \textbf{Method} &\makecell{\textbf{Quality} \\ \textbf{Score}} &\makecell{\textbf{Semantic} \\ \textbf{Score}} & \makecell{\textbf{Subject} \\ \textbf{Consist.}} & \makecell{\textbf{Background} \\ \textbf{Consist.}} & \makecell{\textbf{Temporal} \\ \textbf{Flickering}} & \makecell{\textbf{Motion} \\ \textbf{Smoothness}} &\makecell{\textbf{Dynamic} \\ \textbf{Degree}} &\makecell{\textbf{Aesthetic} \\ \textbf{Quality}} &\makecell{\textbf{Imaging} \\ \textbf{Quality}} \\
        \midrule
        \multirow{3}{*}{VC2} & SFT & 82.1 & 72.2 & 96.9 & 98.1 & 98.4 & 98.3 & 38.9 & 62.4 & 67.4 \\
        & \textbf{\loss} & \textbf{82.3} & \textbf{73.9} & 97.5 & 98.2 & 98.2 & 98.1 & 41.7 & 62.6 & 67.7  \\
        \cmidrule{2-11}
        & Full & \textbf{82.3} & 73.6 & 96.8 & 97.7 & 98.1 & 97.9 & 45.8 & 62.9 & 67.6   \\
        \midrule
        \multirow{3}{*}{AD} & SFT & 81.0 & 74.7 & 98.2 & 97.8 & 98.1 & 98.3 & 20.8 & 65.2 & 66.5   \\
        & \textbf{\loss} & \textbf{81.3} & \textbf{76.8} & 97.8 & 97.9 & 97.9 & 98.3 & 22.2 & 66.5 & 66.9  \\
        \cmidrule{2-11}
        & Full & \textbf{81.3} & 75.1 & 97.2 & 97.8 & 98.0 & 98.1 & 26.4 & 65.7 & 67.1  \\
        \bottomrule
    \end{tabular}}
    \caption{\textbf{Comparison against SFT across multiple models using the VBench evaluation}. Although our primary focus is a human preference-aligned benchmark, we also assess our model's performance on the VBench test sets, which predominantly utilize rule-based measurements across criteria except for aesthetics. Our approach demonstrates robust performance in both Quality and Semantics, consistently outperforming or matching the baseline models and surpassing the SFT-based distillation across all core criteria.}
    \label{tab:vbench_model_comparison}
\end{table*}

\subsection{Experimental Results}
\label{subsec:experimental_results}
Our experimental results are presented in four parts. First, we compare pruned models trained with our method against both naively pruned and full models across two baselines. Second, we evaluate the effectiveness of our \loss loss against the standard distillation loss, SFT. Third, a user study evaluates how well our method aligns with human preferences compared to SFT and full model. Finally, we provide qualitative examples highlighting the advantages of \ours over alternatives. In addition, we evaluate the effectiveness of \ours under extreme conditions using a step-distilled model and address robustness toward potential bias from relying on single reward model by using other reward model, as detailed in Supplementary Sections~\ref{subsec:step_distill} and~\ref{subsec:reward_model}.

\paragraph{Performance of \ours}
Table~\ref{tab:main_ours_comparison} shows the results of applying our framework \ours to VideoCrafter2 and Animatediff, with targeted properties at each stage highlighted in yellow. In most cases, \ours not only improves the explicitly targeted metric but also enhances other properties. Even though some stages show slight drops in dynamic scores, this reflects a natural trade-off, where higher temporal consistency can moderate motion dynamics.  Importantly, ours maintain strong visual quality and temporal coherence. This aligns with previous studies~\cite{liao2025evaluation, wu2024freeinit, xia2024unictrl}, which report that excessive dynamic motion with low temporal consistency often degrades visual quality. These findings highlight that higher dynamics are not always preferable, and that achieving a balanced trade-off between motion and consistency is essential for generating high-quality videos.

Moreover, our final-stage results match or exceed the full model's performance in all metrics for VideoCrafter2, and show similar improvements for Animatediff. Notably, this is achieved with a 36.2\% parameter reduction and a 21\% TFLOPs drop (9.4~$\rightarrow$~7.4) for VideoCrafter2, and a 67.5\% parameter reduction with a 33\% TFLOPs drop (4.9~$\rightarrow$~3.3) for AnimateDiff. These results indicate that our method preserves the strengths inherent in the pruned models, mitigates prior weaknesses, enabling pruned models to outperform full counterparts in both quality and efficiency.

\begin{table}[t]
    \centering
    \scalebox{0.85}{
    \begin{tabular}{lc|cccc}
        \toprule
        \textbf{Model} & \textbf{Method} & \makecell{\textbf{Visual} \\ \textbf{Quality}} & \makecell{\textbf{Temporal} \\ \textbf{Consist.}} & \makecell{\textbf{Dynamic} \\ \textbf{Degree}} & \makecell{\textbf{Text} \\ \textbf{Align.}} \\
        \midrule
        \multirow{4}{*}{VC2}
        & SFT & \underline{2.628} & \underline{2.613} & 2.724 & 2.505 \\
        & \textbf{ReDPO} & \textbf{2.629} & \textbf{2.617} & \textbf{2.728} & \textbf{2.518} \\
        \cmidrule{2-6}
        & Full & 2.627 & 2.602 & \textbf{2.728} & 2.491 \\
        \midrule
        \multirow{4}{*}{AD}
        & SFT & 2.564 & \textbf{2.515} & 2.679 & 2.477 \\
        & \textbf{ReDPO} & \underline{2.569} & \underline{2.513} & \textbf{2.695} & \textbf{2.496} \\
        \cmidrule{2-6}
        & Full & \textbf{2.575} & 2.505 & \underline{2.684} & 2.486 \\
        \bottomrule
    \end{tabular}}
    \caption{\textbf{Comparison against SFT across multiple models on Videoscore}. \textbf{Bold} stands for first place, \underline{underline} stands for second place. Our approach achieves at least second place in all criteria, surpassing both SFT-based distillation and the baseline.}
    \label{tab:qantitative_distillation_comparison}
\end{table}

\paragraph{Comparison with SFT.}
For fair comparison with SFT, we retained our pruning strategy, data curation procedure, and iterative online framework unchanged, replacing only our proposed \loss with the SFT loss. Tables \ref{tab:qantitative_distillation_comparison} and~\ref{tab:vbench_model_comparison} present results comparing SFT-based distillation with full models on VideoScore and VBench test sets, respectively.

Compared to SFT, \loss consistently outperforms across most metrics on both VideoCrafter2 and Animatediff. This is due to fundamental limitation of SFT, which drives distilled models toward averaged predictions under reduced capacity, resulting in blurry outputs and weaker motion dynamics. Such averaging adversely impacts text alignment and overall visual quality. Furthermore, since SFT explicitly attempts to replicate the full model's behavior, it inadvertently reduces properties that pruned models originally excelled, causing unnecessary performance deterioration. In contrast, our method explicitly targets degraded properties and allocates capacity more effectively, guiding the student to focus on essential aspects, resulting in consistently better performance than SFT. 

\begin{figure}[t] \centering
    \includegraphics[width=0.48\textwidth]{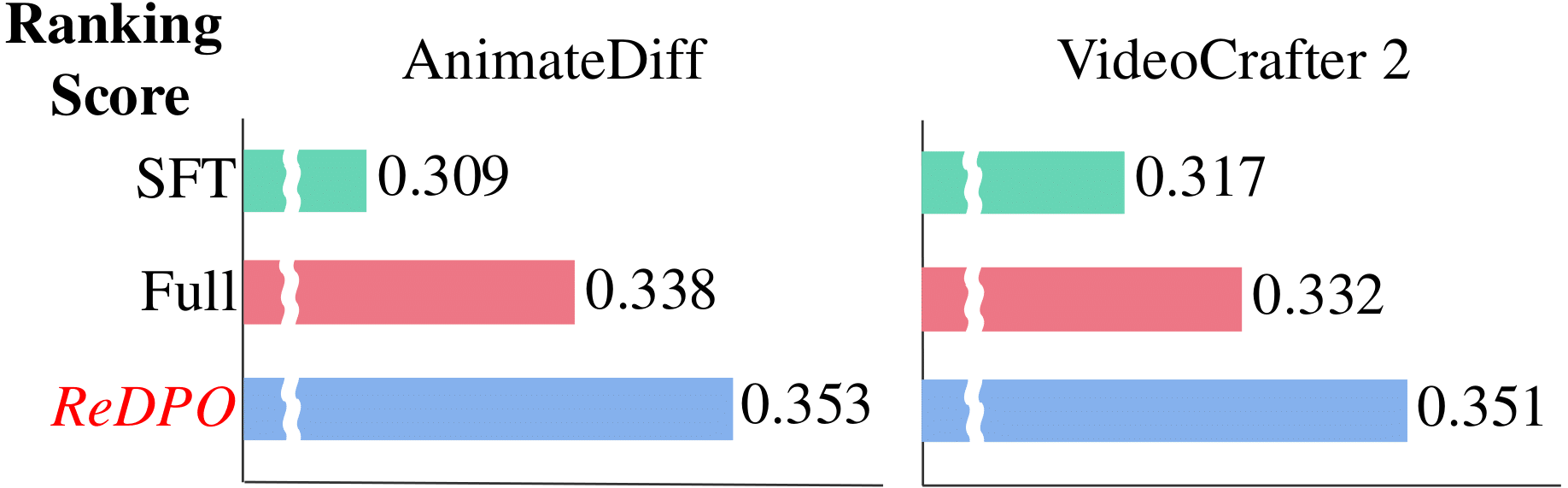}
    \caption{\textbf{Results of User Study}. The results demonstrate that ours outperforms other baselines, indicating that \loss effectively distills the underperforming dimensions from the full model.} \label{fig:userstudy}
\end{figure}

\begin{figure*}[h] \centering
    \includegraphics[width=\textwidth]{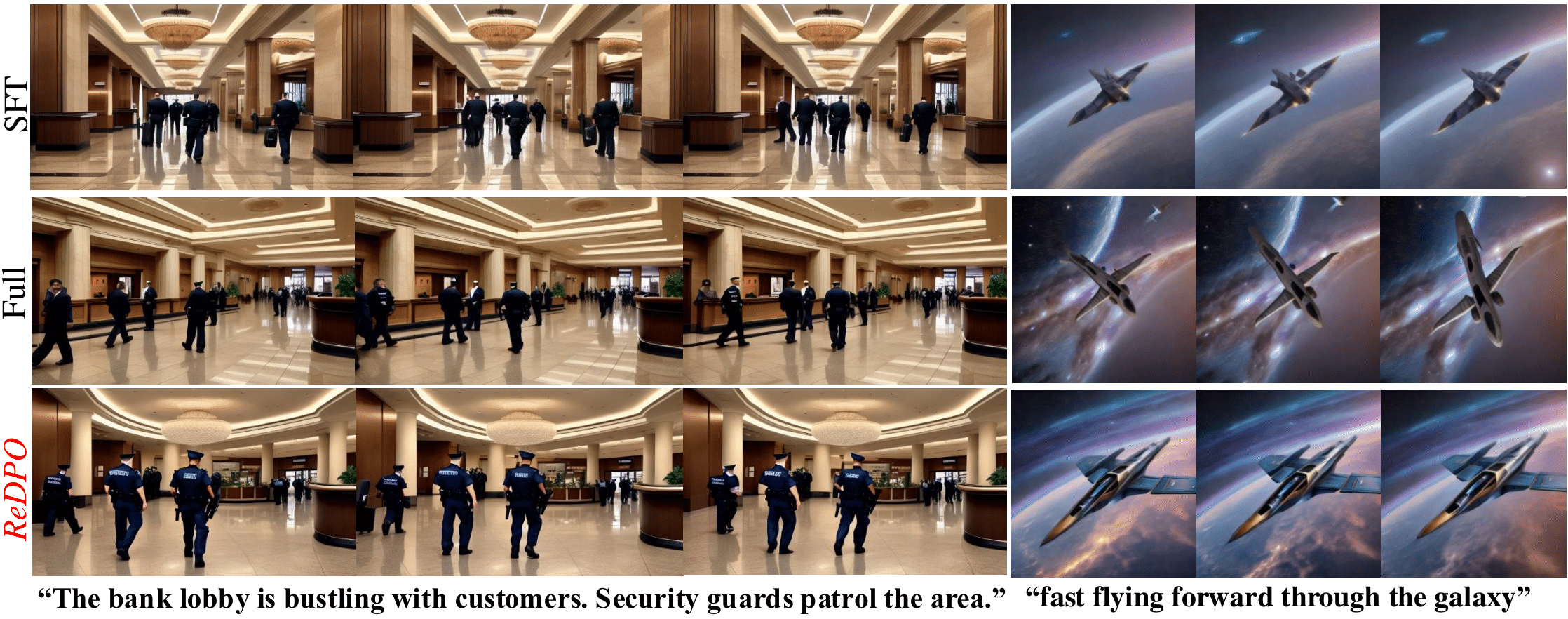}
    \caption{\textbf{Quantitative results of videos with VideoCrafter2 (left) and AnimateDiff (right) using the full and pruned model with different distillation methods}. On the left, only the V.I.P.-trained model successfully generates a security guard, aligning with the prompt. On the right, the \ours trained pruned AnimateDiff achieves the highest visual quality and consistency, whereas other models produce colorless spaceship and blurry outputs.} \label{fig:figure3}
\end{figure*}

\paragraph{User study.}
We also present the user study results in Figure~\ref{fig:userstudy}, demonstrating that \ours significantly outperforms both Full and SFT in terms of overall preference. This results demonstrate that the model trained with our \loss also effectively aligns with human preferences.
Further details are provided in Supplementary Section~\ref{subsec:userstudy_details}.

\paragraph{Visualization.}
As shown in Figure~\ref{fig:figure3}, the left illustrates results from VideoCrafter2 and the right from AnimateDiff. In the left side of Figure~\ref{fig:figure3}, only our \ours framework enables the model to generate a video including security guard, properly reflecting the given prompt while others fail to depict the intended concept. On the right side, AnimateDiff trained with \loss produces the highest visual quality and consistency. The spaceship retains detailed structure, and the background appears more vibrant. In contrast, SFT yields colorless, blurry frames with inconsistent motion where the tail of the spaceship changing across frames. The full model, while more colorful than SFT, also exhibits distortions in the spaceship shape and temporal inconsistency. Despite reduced parameters, \ours framework enables high-quality generation through effective distillation. Further qualitative results can be found in Supplementary.

\begin{table}[t]
    \centering
    \scalebox{0.8}{
    \begin{tabular}{lccccc}
        \toprule
        \textbf{Model} & \textbf{Method} & \makecell{\textbf{Visual} \\ \textbf{Quality}} & \makecell{\textbf{Temporal} \\ \textbf{Consist.}} & \makecell{\textbf{Dynamic} \\ \textbf{Degree}} & \makecell{\textbf{Text} \\ \textbf{Align.}} \\
        \midrule
        \multirow{3}{*}{VC2} & w/o SFT & 2.625 & 2.583 & \textbf{2.729} & 2.471 \\
        & w/o online & 2.626 & 2.603 & 2.719 & 2.483 \\
        & \textbf{\ours} & \textbf{2.629} & \textbf{2.617} & 2.728 & \textbf{2.518} \\
        \midrule
        \multirow{3}{*}{AD} & w/o SFT & 2.563 & 2.437 & \textbf{2.744} & 2.480 \\
        & w/o online & 2.564 & 2.506 & 2.670 & \textbf{2.498} \\
        & \textbf{\ours} & \textbf{2.569} & \textbf{2.513} & 2.695 & 2.496 \\
        \bottomrule
    \end{tabular}}
    \caption{\textbf{Results of ablation study}. The model with our full \ours outperforms other methods, demonstrating the effectiveness of \loss and iterative training.}
    \label{tab:videoscore_ablation}
\end{table}

\subsection{Ablation Study}
\label{subsec:ablation}
Table~\ref{tab:videoscore_ablation} presents the ablation study results on the impact of SFT implementation in our ReDPO loss and the effectiveness of our \ours framework for the online setting. For SFT ablation, we kept the entire framework unchanged and modified only the loss function, and in the offline ablation, we removed all modules at once and trained using \loss. 

When SFT is removed from \loss, we observe a performance drop across nearly all properties in Table~\ref{tab:videoscore_ablation}, confirming that SFT plays a crucial role in maintaining performance. Although the dynamic degree is higher than ours, the drop in consistency indicates that poor motion quality videos are generated. As mentioned earlier, SFT is incorporated as DPO by directly maximizing the relative likelihood of preferred versus losing responses, which can paradoxically degrade the absolute quality of preferred outputs. Therefore, SFT helps explicitly constrain the preference probability to be higher, ensuring high-quality generations. 

Compared to offline setting, our method achieves superior performance across most properties. This highlights the effectiveness of progressively pruning modules and reducing capacity while analyzing degraded properties, rather than dropping the model's capacity all at once. Also, by iteratively generating datasets from pruned models in an on-policy manner, the model enables self-reflection, better aligning with the full model’s distribution and produce high-quality videos. Details and additional evaluations are provided in Supplementary Section~\ref{sec:sup_additional_exp}.

\section{Conclusion}
In this work, we introduced \loss, a novel distillation loss for diffusion models, and \ours, a new framework that integrates \loss into an online step-by-step distillation process. To overcome the limitations of conventional SFT-based distillation, we leverage preference learning through DPO to explicitly guide the model toward targeted property. To address DPO's tendency toward over-optimization, we incorporate SFT-based regularization term, resulting in more stable and effective training. Our \loss consistently outperform SFT in distillation quality, and when combined with our iterative online step-by-step distillation process, enables pruned model to achieve performance comparable to, and in some cases surpassing, the full model, while significantly reducing the number of parameters. This highlights the effectiveness of our approach in enhancing efficiency without compromising generative quality.

\section*{Acknowledgements}

This work was supported by an IITP grant funded by the Korean Government (MSIT) (No. RS-2020-II201361 , Artificial Intelligence Graduate School Program (Yonsei University)) and Institute of Information \& Communications Technology Planning \& Evaluation (IITP) grants funded by the Korea government (MSIT) (No. RS-2024-00457882, AI Research Hub Project) and Culture, Sports and Tourism R\&D Program through the Korea Creative Content Agency grant funded by the Ministry of Culture, Sports and Tourism in 2024 (Project Name:Development of multimodal UX evaluation platform technology for XR spatial responsive content optimization, Project Number: RS-2024-00361757) and Samsung Electronics Co., Ltd(No. IO240424-09660-01).

{
    \small
    \bibliographystyle{ieeenat_fullname}
    \bibliography{main}

\begin{thebibliography}{66}
\providecommand{\natexlab}[1]{#1}
\providecommand{\url}[1]{\texttt{#1}}
\expandafter\ifx\csname urlstyle\endcsname\relax
  \providecommand{\doi}[1]{doi: #1}\else
  \providecommand{\doi}{doi: \begingroup \urlstyle{rm}\Url}\fi

\bibitem[Azar et~al.(2024)Azar, Guo, Piot, Munos, Rowland, Valko, and Calandriello]{azar2024general}
Mohammad~Gheshlaghi Azar, Zhaohan~Daniel Guo, Bilal Piot, Remi Munos, Mark Rowland, Michal Valko, and Daniele Calandriello.
\newblock A general theoretical paradigm to understand learning from human preferences.
\newblock In \emph{International Conference on Artificial Intelligence and Statistics}, pages 4447--4455. PMLR, 2024.

\bibitem[Bain et~al.(2021)Bain, Nagrani, Varol, and Zisserman]{bain2021frozen}
Max Bain, Arsha Nagrani, G{\"u}l Varol, and Andrew Zisserman.
\newblock Frozen in time: A joint video and image encoder for end-to-end retrieval.
\newblock In \emph{Proceedings of the IEEE/CVF international conference on computer vision}, pages 1728--1738, 2021.

\bibitem[Blattmann et~al.(2023{\natexlab{a}})Blattmann, Dockhorn, Kulal, Mendelevitch, Kilian, Lorenz, Levi, English, Voleti, Letts, et~al.]{blattmann2023stablevideodiffusion}
Andreas Blattmann, Tim Dockhorn, Sumith Kulal, Daniel Mendelevitch, Maciej Kilian, Dominik Lorenz, Yam Levi, Zion English, Vikram Voleti, Adam Letts, et~al.
\newblock Stable video diffusion: Scaling latent video diffusion models to large datasets.
\newblock \emph{arXiv preprint arXiv:2311.15127}, 2023{\natexlab{a}}.

\bibitem[Blattmann et~al.(2023{\natexlab{b}})Blattmann, Rombach, Ling, Dockhorn, Kim, Fidler, and Kreis]{blattmann2023videoldm}
Andreas Blattmann, Robin Rombach, Huan Ling, Tim Dockhorn, Seung~Wook Kim, Sanja Fidler, and Karsten Kreis.
\newblock Align your latents: High-resolution video synthesis with latent diffusion models.
\newblock In \emph{Proceedings of the IEEE/CVF conference on computer vision and pattern recognition}, pages 22563--22575, 2023{\natexlab{b}}.

\bibitem[Blau and Michaeli(2018)]{mse_bad_perception_distortion}
Yochai Blau and Tomer Michaeli.
\newblock The perception-distortion tradeoff.
\newblock In \emph{Proceedings of the IEEE conference on computer vision and pattern recognition}, pages 6228--6237, 2018.

\bibitem[Castells et~al.(2024)Castells, Song, Piao, Choi, Kim, Yim, Lee, Kim, and Kim]{castells2024edgefusion}
Thibault Castells, Hyoung-Kyu Song, Tairen Piao, Shinkook Choi, Bo-Kyeong Kim, Hanyoung Yim, Changgwun Lee, Jae~Gon Kim, and Tae-Ho Kim.
\newblock Edgefusion: on-device text-to-image generation.
\newblock \emph{arXiv preprint arXiv:2404.11925}, 2024.

\bibitem[Chen et~al.(2024{\natexlab{a}})Chen, Zhang, Cun, Xia, Wang, Weng, and Shan]{chen2024videocrafter2}
Haoxin Chen, Yong Zhang, Xiaodong Cun, Menghan Xia, Xintao Wang, Chao Weng, and Ying Shan.
\newblock Videocrafter2: Overcoming data limitations for high-quality video diffusion models.
\newblock In \emph{Proceedings of the IEEE/CVF Conference on Computer Vision and Pattern Recognition}, pages 7310--7320, 2024{\natexlab{a}}.

\bibitem[Chen et~al.(2024{\natexlab{b}})Chen, Deng, Yuan, Ji, and Gu]{chen2024self}
Zixiang Chen, Yihe Deng, Huizhuo Yuan, Kaixuan Ji, and Quanquan Gu.
\newblock Self-play fine-tuning converts weak language models to strong language models.
\newblock \emph{arXiv preprint arXiv:2401.01335}, 2024{\natexlab{b}}.

\bibitem[Cheng et~al.(2025)Cheng, Lyu, Gu, Liu, Xu, Lu, Teng, Yang, Dong, Tang, et~al.]{cheng2025vpo}
Jiale Cheng, Ruiliang Lyu, Xiaotao Gu, Xiao Liu, Jiazheng Xu, Yida Lu, Jiayan Teng, Zhuoyi Yang, Yuxiao Dong, Jie Tang, et~al.
\newblock Vpo: Aligning text-to-video generation models with prompt optimization.
\newblock \emph{arXiv preprint arXiv:2503.20491}, 2025.

\bibitem[Dong et~al.(2024)Dong, Xiong, Pang, Wang, Zhao, Zhou, Jiang, Sahoo, Xiong, and Zhang]{dong2024rlhf}
Hanze Dong, Wei Xiong, Bo Pang, Haoxiang Wang, Han Zhao, Yingbo Zhou, Nan Jiang, Doyen Sahoo, Caiming Xiong, and Tong Zhang.
\newblock Rlhf workflow: From reward modeling to online rlhf.
\newblock \emph{arXiv preprint arXiv:2405.07863}, 2024.

\bibitem[Ethayarajh et~al.(2024)Ethayarajh, Xu, Muennighoff, Jurafsky, and Kiela]{ethayarajh2024kto}
Kawin Ethayarajh, Winnie Xu, Niklas Muennighoff, Dan Jurafsky, and Douwe Kiela.
\newblock Kto: Model alignment as prospect theoretic optimization.
\newblock \emph{arXiv preprint arXiv:2402.01306}, 2024.

\bibitem[Fisch et~al.(2024)Fisch, Eisenstein, Zayats, Agarwal, Beirami, Nagpal, Shaw, and Berant]{fisch2024robust}
Adam Fisch, Jacob Eisenstein, Vicky Zayats, Alekh Agarwal, Ahmad Beirami, Chirag Nagpal, Pete Shaw, and Jonathan Berant.
\newblock Robust preference optimization through reward model distillation.
\newblock \emph{arXiv preprint arXiv:2405.19316}, 2024.

\bibitem[Freirich et~al.(2021)Freirich, Michaeli, and Meir]{mse_bad_3_wasserstein}
Dror Freirich, Tomer Michaeli, and Ron Meir.
\newblock A theory of the distortion-perception tradeoff in wasserstein space.
\newblock \emph{Advances in Neural Information Processing Systems}, 34:\penalty0 25661--25672, 2021.

\bibitem[Gao et~al.(2025)Gao, Gao, Wu, Zhou, Qiao, Niu, Chen, and Wang]{gao2025devil}
Bingjie Gao, Xinyu Gao, Xiaoxue Wu, Yujie Zhou, Yu Qiao, Li Niu, Xinyuan Chen, and Yaohui Wang.
\newblock The devil is in the prompts: Retrieval-augmented prompt optimization for text-to-video generation.
\newblock In \emph{Proceedings of the Computer Vision and Pattern Recognition Conference}, pages 3173--3183, 2025.

\bibitem[Goodfellow et~al.(2020)Goodfellow, Pouget-Abadie, Mirza, Xu, Warde-Farley, Ozair, Courville, and Bengio]{goodfellow2020generative}
Ian Goodfellow, Jean Pouget-Abadie, Mehdi Mirza, Bing Xu, David Warde-Farley, Sherjil Ozair, Aaron Courville, and Yoshua Bengio.
\newblock Generative adversarial networks.
\newblock \emph{Communications of the ACM}, 63\penalty0 (11):\penalty0 139--144, 2020.

\bibitem[Guo et~al.(2024)Guo, Zhang, Liu, Liu, Khalman, Llinares, Rame, Mesnard, Zhao, Piot, et~al.]{guo2024direct}
Shangmin Guo, Biao Zhang, Tianlin Liu, Tianqi Liu, Misha Khalman, Felipe Llinares, Alexandre Rame, Thomas Mesnard, Yao Zhao, Bilal Piot, et~al.
\newblock Direct language model alignment from online ai feedback.
\newblock \emph{arXiv preprint arXiv:2402.04792}, 2024.

\bibitem[Guo et~al.(2023)Guo, Yang, Rao, Liang, Wang, Qiao, Agrawala, Lin, and Dai]{guo2023animatediff}
Yuwei Guo, Ceyuan Yang, Anyi Rao, Zhengyang Liang, Yaohui Wang, Yu Qiao, Maneesh Agrawala, Dahua Lin, and Bo Dai.
\newblock Animatediff: Animate your personalized text-to-image diffusion models without specific tuning.
\newblock \emph{arXiv preprint arXiv:2307.04725}, 2023.

\bibitem[He et~al.(2024)He, Jiang, Zhang, Ku, Soni, Siu, Chen, Chandra, Jiang, Arulraj, et~al.]{he2024videoscore}
Xuan He, Dongfu Jiang, Ge Zhang, Max Ku, Achint Soni, Sherman Siu, Haonan Chen, Abhranil Chandra, Ziyan Jiang, Aaran Arulraj, et~al.
\newblock Videoscore: Building automatic metrics to simulate fine-grained human feedback for video generation.
\newblock \emph{arXiv preprint arXiv:2406.15252}, 2024.

\bibitem[He et~al.(2017)He, Zhang, and Sun]{he2017channel}
Yihui He, Xiangyu Zhang, and Jian Sun.
\newblock Channel pruning for accelerating very deep neural networks.
\newblock In \emph{Proceedings of the IEEE international conference on computer vision}, pages 1389--1397, 2017.

\bibitem[Hinton et~al.(2015)Hinton, Vinyals, and Dean]{hinton2015distilling}
Geoffrey Hinton, Oriol Vinyals, and Jeff Dean.
\newblock Distilling the knowledge in a neural network.
\newblock \emph{arXiv preprint arXiv:1503.02531}, 2015.

\bibitem[Ho et~al.(2020)Ho, Jain, and Abbeel]{ho2020denoising}
Jonathan Ho, Ajay Jain, and Pieter Abbeel.
\newblock Denoising diffusion probabilistic models.
\newblock \emph{Advances in neural information processing systems}, 33:\penalty0 6840--6851, 2020.

\bibitem[Hong et~al.(2024)Hong, Lee, and Thorne]{hong2024orpo}
Jiwoo Hong, Noah Lee, and James Thorne.
\newblock Orpo: Monolithic preference optimization without reference model.
\newblock \emph{arXiv preprint arXiv:2403.07691}, 2024.

\bibitem[Huang et~al.(2024)Huang, He, Yu, Zhang, Si, Jiang, Zhang, Wu, Jin, Chanpaisit, et~al.]{huang2024vbench}
Ziqi Huang, Yinan He, Jiashuo Yu, Fan Zhang, Chenyang Si, Yuming Jiang, Yuanhan Zhang, Tianxing Wu, Qingyang Jin, Nattapol Chanpaisit, et~al.
\newblock Vbench: Comprehensive benchmark suite for video generative models.
\newblock In \emph{Proceedings of the IEEE/CVF Conference on Computer Vision and Pattern Recognition}, pages 21807--21818, 2024.

\bibitem[Ivison et~al.(2025)Ivison, Wang, Liu, Wu, Pyatkin, Lambert, Smith, Choi, and Hajishirzi]{ivison2025unpacking}
Hamish Ivison, Yizhong Wang, Jiacheng Liu, Zeqiu Wu, Valentina Pyatkin, Nathan Lambert, Noah~A Smith, Yejin Choi, and Hanna Hajishirzi.
\newblock Unpacking dpo and ppo: Disentangling best practices for learning from preference feedback.
\newblock \emph{Advances in neural information processing systems}, 37:\penalty0 36602--36633, 2025.

\bibitem[Jiang et~al.(2025)Jiang, Wu, Zhang, Guan, and Chen]{jiang2025huvidpo}
Lifan Jiang, Boxi Wu, Jiahui Zhang, Xiaotong Guan, and Shuang Chen.
\newblock Huvidpo: Enhancing video generation through direct preference optimization for human-centric alignment.
\newblock \emph{arXiv preprint arXiv:2502.01690}, 2025.

\bibitem[Kim et~al.(2024)Kim, Song, Castells, and Choi]{kim2024bk}
Bo-Kyeong Kim, Hyoung-Kyu Song, Thibault Castells, and Shinkook Choi.
\newblock Bk-sdm: A lightweight, fast, and cheap version of stable diffusion.
\newblock In \emph{European Conference on Computer Vision}, pages 381--399. Springer, 2024.

\bibitem[Kingma et~al.(2013)Kingma, Welling, et~al.]{kingma2013auto}
Diederik~P Kingma, Max Welling, et~al.
\newblock Auto-encoding variational bayes, 2013.

\bibitem[Lagunas et~al.(2021)Lagunas, Charlaix, Sanh, and Rush]{lagunas2021block}
Fran{\c{c}}ois Lagunas, Ella Charlaix, Victor Sanh, and Alexander~M Rush.
\newblock Block pruning for faster transformers.
\newblock \emph{arXiv preprint arXiv:2109.04838}, 2021.

\bibitem[Li et~al.(2024)Li, Feng, Fu, Wang, Basu, Chen, and Wang]{li2024t2v}
Jiachen Li, Weixi Feng, Tsu-Jui Fu, Xinyi Wang, Sugato Basu, Wenhu Chen, and William~Yang Wang.
\newblock T2v-turbo: Breaking the quality bottleneck of video consistency model with mixed reward feedback.
\newblock \emph{arXiv preprint arXiv:2405.18750}, 2024.

\bibitem[Li et~al.(2023)Li, Wang, Jin, Hu, Chemerys, Fu, Wang, Tulyakov, and Ren]{li2023snapfusion}
Yanyu Li, Huan Wang, Qing Jin, Ju Hu, Pavlo Chemerys, Yun Fu, Yanzhi Wang, Sergey Tulyakov, and Jian Ren.
\newblock Snapfusion: Text-to-image diffusion model on mobile devices within two seconds.
\newblock \emph{Advances in Neural Information Processing Systems}, 36:\penalty0 20662--20678, 2023.

\bibitem[Liao et~al.(2024)Liao, Ye, Zuo, Wan, Wang, Zhao, Wang, Zhang, et~al.]{liao2024evaluation}
Mingxiang Liao, Qixiang Ye, Wangmeng Zuo, Fang Wan, Tianyu Wang, Yuzhong Zhao, Jingdong Wang, Xinyu Zhang, et~al.
\newblock Evaluation of text-to-video generation models: A dynamics perspective.
\newblock \emph{Advances in Neural Information Processing Systems}, 37:\penalty0 109790--109816, 2024.

\bibitem[Liao et~al.(2025)Liao, Ye, Zuo, Wan, Wang, Zhao, Wang, Zhang, et~al.]{liao2025evaluation}
Mingxiang Liao, Qixiang Ye, Wangmeng Zuo, Fang Wan, Tianyu Wang, Yuzhong Zhao, Jingdong Wang, Xinyu Zhang, et~al.
\newblock Evaluation of text-to-video generation models: A dynamics perspective.
\newblock \emph{Advances in Neural Information Processing Systems}, 37:\penalty0 109790--109816, 2025.

\bibitem[Lin and Yang(2024)]{lin2024animatediff_lightning}
Shanchuan Lin and Xiao Yang.
\newblock Animatediff-lightning: Cross-model diffusion distillation.
\newblock \emph{arXiv preprint arXiv:2403.12706}, 2024.

\bibitem[Lin et~al.(2024{\natexlab{a}})Lin, Wang, and Yang]{lin2024sdxl}
Shanchuan Lin, Anran Wang, and Xiao Yang.
\newblock Sdxl-lightning: Progressive adversarial diffusion distillation.
\newblock \emph{arXiv preprint arXiv:2402.13929}, 2024{\natexlab{a}}.

\bibitem[Lin et~al.(2024{\natexlab{b}})Lin, Wang, and Yang]{lin2024sdxl_lightning}
Shanchuan Lin, Anran Wang, and Xiao Yang.
\newblock Sdxl-lightning: Progressive adversarial diffusion distillation.
\newblock \emph{arXiv preprint arXiv:2402.13929}, 2024{\natexlab{b}}.

\bibitem[Liu et~al.(2025)Liu, Liu, Liang, Yuan, Liu, Zheng, Wu, Wang, Qin, Xia, et~al.]{liu2025improving}
Jie Liu, Gongye Liu, Jiajun Liang, Ziyang Yuan, Xiaokun Liu, Mingwu Zheng, Xiele Wu, Qiulin Wang, Wenyu Qin, Menghan Xia, et~al.
\newblock Improving video generation with human feedback.
\newblock \emph{arXiv preprint arXiv:2501.13918}, 2025.

\bibitem[Liu et~al.(2024{\natexlab{a}})Liu, Wu, Ziqiang, Wei, He, Pi, and Chen]{liu2024videodpo}
Runtao Liu, Haoyu Wu, Zheng Ziqiang, Chen Wei, Yingqing He, Renjie Pi, and Qifeng Chen.
\newblock Videodpo: Omni-preference alignment for video diffusion generation.
\newblock \emph{arXiv preprint arXiv:2412.14167}, 2024{\natexlab{a}}.

\bibitem[Liu et~al.(2017)Liu, Li, Shen, Huang, Yan, and Zhang]{liu2017learning}
Zhuang Liu, Jianguo Li, Zhiqiang Shen, Gao Huang, Shoumeng Yan, and Changshui Zhang.
\newblock Learning efficient convolutional networks through network slimming.
\newblock In \emph{Proceedings of the IEEE international conference on computer vision}, pages 2736--2744, 2017.

\bibitem[Liu et~al.(2024{\natexlab{b}})Liu, Lu, Zhang, Liu, Guo, Yang, Blanchet, and Wang]{liu2024provably}
Zhihan Liu, Miao Lu, Shenao Zhang, Boyi Liu, Hongyi Guo, Yingxiang Yang, Jose Blanchet, and Zhaoran Wang.
\newblock Provably mitigating overoptimization in rlhf: Your sft loss is implicitly an adversarial regularizer.
\newblock \emph{arXiv preprint arXiv:2405.16436}, 2024{\natexlab{b}}.

\bibitem[Meng et~al.(2025)Meng, Xia, and Chen]{meng2025simpo}
Yu Meng, Mengzhou Xia, and Danqi Chen.
\newblock Simpo: Simple preference optimization with a reference-free reward.
\newblock \emph{Advances in Neural Information Processing Systems}, 37:\penalty0 124198--124235, 2025.

\bibitem[Ohayon et~al.(2024)Ohayon, Michaeli, and Elad]{mse_bad_2_pmrf}
Guy Ohayon, Tomer Michaeli, and Michael Elad.
\newblock Posterior-mean rectified flow: Towards minimum mse photo-realistic image restoration.
\newblock \emph{arXiv preprint arXiv:2410.00418}, 2024.

\bibitem[Ouyang et~al.(2022)Ouyang, Wu, Jiang, Almeida, Wainwright, Mishkin, Zhang, Agarwal, Slama, Ray, et~al.]{ouyang2022training}
Long Ouyang, Jeffrey Wu, Xu Jiang, Diogo Almeida, Carroll Wainwright, Pamela Mishkin, Chong Zhang, Sandhini Agarwal, Katarina Slama, Alex Ray, et~al.
\newblock Training language models to follow instructions with human feedback.
\newblock \emph{Advances in neural information processing systems}, 35:\penalty0 27730--27744, 2022.

\bibitem[Pal et~al.(2024)Pal, Karkhanis, Dooley, Roberts, Naidu, and White]{pal2024smaug}
Arka Pal, Deep Karkhanis, Samuel Dooley, Manley Roberts, Siddartha Naidu, and Colin White.
\newblock Smaug: Fixing failure modes of preference optimisation with dpo-positive.
\newblock \emph{arXiv preprint arXiv:2402.13228}, 2024.

\bibitem[Podell et~al.(2023)Podell, English, Lacey, Blattmann, Dockhorn, M{\"u}ller, Penna, and Rombach]{podell2023sdxl}
Dustin Podell, Zion English, Kyle Lacey, Andreas Blattmann, Tim Dockhorn, Jonas M{\"u}ller, Joe Penna, and Robin Rombach.
\newblock Sdxl: Improving latent diffusion models for high-resolution image synthesis.
\newblock \emph{arXiv preprint arXiv:2307.01952}, 2023.

\bibitem[Rafailov et~al.(2023)Rafailov, Sharma, Mitchell, Manning, Ermon, and Finn]{rafailov2023direct}
Rafael Rafailov, Archit Sharma, Eric Mitchell, Christopher~D Manning, Stefano Ermon, and Chelsea Finn.
\newblock Direct preference optimization: Your language model is secretly a reward model.
\newblock \emph{Advances in Neural Information Processing Systems}, 36:\penalty0 53728--53741, 2023.

\bibitem[Ren et~al.(2025)Ren, Zhang, Liu, Zhang, and Tian]{ren2025refining}
Jie Ren, Yuhang Zhang, Dongrui Liu, Xiaopeng Zhang, and Qi Tian.
\newblock Refining alignment framework for diffusion models with intermediate-step preference ranking.
\newblock \emph{arXiv preprint arXiv:2502.01667}, 2025.

\bibitem[Rombach et~al.(2022{\natexlab{a}})Rombach, Blattmann, Lorenz, Esser, and Ommer]{rombach2022high}
Robin Rombach, Andreas Blattmann, Dominik Lorenz, Patrick Esser, and Bj{\"o}rn Ommer.
\newblock High-resolution image synthesis with latent diffusion models.
\newblock In \emph{Proceedings of the IEEE/CVF conference on computer vision and pattern recognition}, pages 10684--10695, 2022{\natexlab{a}}.

\bibitem[Rombach et~al.(2022{\natexlab{b}})Rombach, Blattmann, Lorenz, Esser, and Ommer]{rombach2022ldm}
Robin Rombach, Andreas Blattmann, Dominik Lorenz, Patrick Esser, and Bj{\"o}rn Ommer.
\newblock High-resolution image synthesis with latent diffusion models.
\newblock In \emph{Proceedings of the IEEE/CVF conference on computer vision and pattern recognition}, pages 10684--10695, 2022{\natexlab{b}}.

\bibitem[Schulman et~al.(2017)Schulman, Wolski, Dhariwal, Radford, and Klimov]{schulman2017proximal}
John Schulman, Filip Wolski, Prafulla Dhariwal, Alec Radford, and Oleg Klimov.
\newblock Proximal policy optimization algorithms.
\newblock \emph{arXiv preprint arXiv:1707.06347}, 2017.

\bibitem[Su et~al.(2024)Su, Liu, Gao, and Song]{su2024f3}
Sitong Su, Jianzhi Liu, Lianli Gao, and Jingkuan Song.
\newblock F$^3$-pruning: A training-free and generalized pruning strategy towards faster and finer text-to-video synthesis.
\newblock In \emph{Proceedings of the AAAI Conference on Artificial Intelligence}, pages 4961--4969, 2024.

\bibitem[Sun et~al.(2025)Sun, Hou, Di, Yang, Ma, and Cui]{sun2025unicp}
Wenzhang Sun, Qirui Hou, Donglin Di, Jiahui Yang, Yongjia Ma, and Jianxun Cui.
\newblock Unicp: A unified caching and pruning framework for efficient video generation.
\newblock \emph{arXiv preprint arXiv:2502.04393}, 2025.

\bibitem[Tajwar et~al.(2024)Tajwar, Singh, Sharma, Rafailov, Schneider, Xie, Ermon, Finn, and Kumar]{tajwar2024preference}
Fahim Tajwar, Anikait Singh, Archit Sharma, Rafael Rafailov, Jeff Schneider, Tengyang Xie, Stefano Ermon, Chelsea Finn, and Aviral Kumar.
\newblock Preference fine-tuning of llms should leverage suboptimal, on-policy data.
\newblock \emph{arXiv preprint arXiv:2404.14367}, 2024.

\bibitem[Tang et~al.(2024)Tang, Guo, Zheng, Calandriello, Cao, Tarassov, Munos, Pires, Valko, Cheng, et~al.]{tang2024understanding}
Yunhao Tang, Daniel~Zhaohan Guo, Zeyu Zheng, Daniele Calandriello, Yuan Cao, Eugene Tarassov, R{\'e}mi Munos, Bernardo~{\'A}vila Pires, Michal Valko, Yong Cheng, et~al.
\newblock Understanding the performance gap between online and offline alignment algorithms.
\newblock \emph{arXiv preprint arXiv:2405.08448}, 2024.

\bibitem[Wallace et~al.(2024)Wallace, Dang, Rafailov, Zhou, Lou, Purushwalkam, Ermon, Xiong, Joty, and Naik]{wallace2024diffusion}
Bram Wallace, Meihua Dang, Rafael Rafailov, Linqi Zhou, Aaron Lou, Senthil Purushwalkam, Stefano Ermon, Caiming Xiong, Shafiq Joty, and Nikhil Naik.
\newblock Diffusion model alignment using direct preference optimization.
\newblock In \emph{Proceedings of the IEEE/CVF Conference on Computer Vision and Pattern Recognition}, pages 8228--8238, 2024.

\bibitem[Wan et~al.(2025)Wan, Wang, Ai, Wen, Mao, Xie, Chen, Yu, Zhao, Yang, et~al.]{wan2025wan}
Team Wan, Ang Wang, Baole Ai, Bin Wen, Chaojie Mao, Chen-Wei Xie, Di Chen, Feiwu Yu, Haiming Zhao, Jianxiao Yang, et~al.
\newblock Wan: Open and advanced large-scale video generative models.
\newblock \emph{arXiv preprint arXiv:2503.20314}, 2025.

\bibitem[Wu et~al.(2024{\natexlab{a}})Wu, Si, Jiang, Huang, and Liu]{wu2024freeinit}
Tianxing Wu, Chenyang Si, Yuming Jiang, Ziqi Huang, and Ziwei Liu.
\newblock Freeinit: Bridging initialization gap in video diffusion models.
\newblock In \emph{European Conference on Computer Vision}, pages 378--394. Springer, 2024{\natexlab{a}}.

\bibitem[Wu et~al.(2024{\natexlab{b}})Wu, Wang, Chen, and Xu]{wu2024individual}
Yiming Wu, Huan Wang, Zhenghao Chen, and Dong Xu.
\newblock Individual content and motion dynamics preserved pruning for video diffusion models.
\newblock \emph{arXiv preprint arXiv:2411.18375}, 2024{\natexlab{b}}.

\bibitem[Wu et~al.(2024{\natexlab{c}})Wu, Zhang, Li, Xu, Kag, Sui, Coskun, Ma, Lebedev, Hu, et~al.]{wu2024snapgen}
Yushu Wu, Zhixing Zhang, Yanyu Li, Yanwu Xu, Anil Kag, Yang Sui, Huseyin Coskun, Ke Ma, Aleksei Lebedev, Ju Hu, et~al.
\newblock Snapgen-v: Generating a five-second video within five seconds on a mobile device.
\newblock \emph{arXiv preprint arXiv:2412.10494}, 2024{\natexlab{c}}.

\bibitem[Xia et~al.(2022)Xia, Zhong, and Chen]{xia2022structured}
Mengzhou Xia, Zexuan Zhong, and Danqi Chen.
\newblock Structured pruning learns compact and accurate models.
\newblock \emph{arXiv preprint arXiv:2204.00408}, 2022.

\bibitem[Xia et~al.(2024)Xia, Chen, and Xu]{xia2024unictrl}
Tian Xia, Xuweiyi Chen, and Sihan Xu.
\newblock Unictrl: Improving the spatiotemporal consistency of text-to-video diffusion models via training-free unified attention control.
\newblock \emph{arXiv preprint arXiv:2403.02332}, 2024.

\bibitem[Xiao et~al.(2021)Xiao, Kreis, and Vahdat]{xiao2021tackling}
Zhisheng Xiao, Karsten Kreis, and Arash Vahdat.
\newblock Tackling the generative learning trilemma with denoising diffusion gans.
\newblock \emph{arXiv preprint arXiv:2112.07804}, 2021.

\bibitem[Yahia et~al.(2024)Yahia, Korzhenkov, Lelekas, Ghodrati, and Habibian]{yahia2024mobile}
Haitam~Ben Yahia, Denis Korzhenkov, Ioannis Lelekas, Amir Ghodrati, and Amirhossein Habibian.
\newblock Mobile video diffusion.
\newblock \emph{arXiv preprint arXiv:2412.07583}, 2024.

\bibitem[Yang et~al.(2024)Yang, Tao, Lyu, Ge, Chen, Shen, Zhu, and Li]{yang2024using}
Kai Yang, Jian Tao, Jiafei Lyu, Chunjiang Ge, Jiaxin Chen, Weihan Shen, Xiaolong Zhu, and Xiu Li.
\newblock Using human feedback to fine-tune diffusion models without any reward model.
\newblock In \emph{Proceedings of the IEEE/CVF Conference on Computer Vision and Pattern Recognition}, pages 8941--8951, 2024.

\bibitem[Zhang et~al.(2024)Zhang, Wu, Chen, Ji, Xiao, Huang, and Han]{zhang2024onlinevpo}
Jiacheng Zhang, Jie Wu, Weifeng Chen, Yatai Ji, Xuefeng Xiao, Weilin Huang, and Kai Han.
\newblock Onlinevpo: Align video diffusion model with online video-centric preference optimization.
\newblock \emph{arXiv preprint arXiv:2412.15159}, 2024.

\bibitem[Zhao et~al.(2024)Zhao, Xu, Xiao, Jia, and Hou]{zhao2024mobilediffusion}
Yang Zhao, Yanwu Xu, Zhisheng Xiao, Haolin Jia, and Tingbo Hou.
\newblock Mobilediffusion: Instant text-to-image generation on mobile devices.
\newblock In \emph{European Conference on Computer Vision}, pages 225--242. Springer, 2024.

\bibitem[Ziegler et~al.(2019)Ziegler, Stiennon, Wu, Brown, Radford, Amodei, Christiano, and Irving]{ziegler2019fine}
Daniel~M Ziegler, Nisan Stiennon, Jeffrey Wu, Tom~B Brown, Alec Radford, Dario Amodei, Paul Christiano, and Geoffrey Irving.
\newblock Fine-tuning language models from human preferences.
\newblock \emph{arXiv preprint arXiv:1909.08593}, 2019.

\end{thebibliography}
}
\clearpage
\setcounter{page}{1}
\maketitlesupplementary
\renewcommand{\thesection}{\Alph{section}}
\setcounter{section}{0} 
\renewcommand{\thetable}{\Alph{table}}
\setcounter{table}{0}
\renewcommand{\thefigure}{\Alph{figure}}
\setcounter{figure}{0}

\definecolor{codebg}{rgb}{0.95, 0.95, 0.95}  %
\lstdefinestyle{mystyle}{
    backgroundcolor=\color{codebg},   %
    basicstyle=\ttfamily\small,       %
    frame=single,                     %
    breaklines=true,                   %
    captionpos=b,                      %
    keywordstyle=\bfseries,            %
    commentstyle=\color{gray},         %
    numbersep=5pt,                     %
    xleftmargin=5pt, xrightmargin=5pt  %
}

In this supplementary material, we provide,
\addcontentsline{toc}{section}{Appendix}
\begin{itemize}
    \item \textbf{A. Limitations}
    \item \textbf{B. Future Research}
    \item \textbf{B. Experimental Details}
    \begin{itemize}
        \item B.1. Pruning algorithm details
        \item B.2. Data curation details
        \item B.3. Evaluation details
        \item B.4. Dynamic degree analysis
        \item B.5. User study details
    \end{itemize}
    \item \textbf{C. Additional Experiments}
    \begin{itemize}
        \item C.1. SFT Weight Experiment
        \item C.2. Further Experiments for VideoCrafter 2
        \item C.3. Experiments for step-distilled model
        \item C.4. Experiments for reward model
    \end{itemize}
    \item \textbf{D. Additional Explanations on Motivation}
    \item \textbf{E. Prompt Filtering}
        \begin{itemize}
        \item E.1 Dynamic Degree
        \item E.2 Visual Quality
        \item E.3 Text Alignment
    \end{itemize}
    \item \textbf{F. Qualitative Results}
\end{itemize}

\section{Limitations}
Since we used VideoScore as the reward model, further advancements in reward modeling could further enhance our performance. As our experiments are conducted only on U-Net models, further research on Transformer-based models could provide additional insights and improvements.

\section{Future Research}
As pruning methods for DiT-based video diffusion model\cite{sun2025unicp} are beginning to emerge, \ours, which is \textit{orthogonal} to any pruning strategy, can be applied to DiT-based models by simply adapting such methodology into the framework. Moreover, while DiT-based models benefit from the scalability of transformer architecture, they are computationally heavy. We believe that \ours could also serve as an effective distillation method by leveraging large, capable DiT models as teachers and smaller models as students (e.g. Wan2.1 13B \& 1.3B)\cite{wan2025wan}, contributing to building practical T2V models.

\section{Experimental details}
\label{sec:sup_experimental_details}
In this section, we report the additional details of training. As for the prompt filtering process, we detail it in Section~\ref{sec:prompt_filter} for better readability.

\subsection{Pruning algorithm details}
\label{sec:sup_pruning_algorithm}
At the pruning stage, we iteratively removed blocks from the model one by one and evaluated each model using VideoScore~\cite{he2024videoscore}. We then sorted the models by total VideoScore and selected four motion module blocks for AnimateDiff and four U-Net blocks for VideoCrafter2 from the top-ranked models at each stage.

However, as previously mentioned, we observed cases where a high total score was misleading—some models achieved a high total score due to extremely high dynamic degree, despite having low consistency. This suggests that motion quality was poor, but the overall score was inflated because the drop in consistency was offset by an unusually high dynamic degree.

To address this issue, we sorted models by both total VideoScore and consistency to ensure that motion quality was properly considered. Finally, our pruning stage concluded by selecting blocks based on the intersection of the highest-ranked models in total VideoScore and the highest-ranked models in consistency.

\begin{algorithm}
\caption{Step-by-Step Pruning Algorithm}
\label{alg:progressive_pruning}
\begin{algorithmic}
    \State \textbf{Input:} Model $M$ with $N$ modules, Benchmark $V_{\text{bench}}$, Number of modules to prune per stage $k$
    \State \textbf{Output:} Pruned and Distilled Model $M_{\text{pruned}}$
    
    \While{pruning not completed}
        \For{$i \gets 1$ to $N$}
            \State Compute $\Delta_i = V_{\text{metric}}(M) - V_{\text{metric}}(M - \text{module } i)$
        \EndFor 
        \State Select $k$ modules in ascending order of $\Delta_i$: $\mathcal{S}=\{ m_1, ..., m_k \}$ 
        \State Prune $\mathcal{S}$ from $M$: $M_{\text{pruned}} \gets M - \mathcal{S}$, \quad $N \gets N - k$
        \State Perform dataset curation and train $M_{\text{pruned}}$
        \State Update $M \gets M_{\text{pruned}}$
    \EndWhile
\end{algorithmic}
\end{algorithm}

\subsection{Data curation details}
\label{sec:data_curation_details}
After pruning four modules, we re-evaluated the model using VideoScore to identify which properties had decreased compared to the full model. Note that, as previously mentioned, removing certain redundant blocks can sometimes improve performance rather than degrade it. If multiple properties exhibited a drop, we selected the property with the largest gap from the full model as the primary target. 

Based on this, we used target-filtered prompts obtained by prompt filtering, and used them to generate prefer-unprefer datasets from the full model and pruned model each. If multiple target properties were selected, we ensured that all selected properties were considered when forming preferred-unpreferred pairs. Moreover, since unpreferred pairs are also part of the training dataset, their quality is crucial. To maintain meaningful learning, we introduced a lower bound for unpreferred pairs, ensuring that their scores remained above $mean - \alpha * std$, and $\alpha$ is fixed at 0.3 in all stages.

Additionally, to prevent unintended learning where other properties dominate the preference learning, we imposed a threshold condition ensuring that the gap in the targeted property (preferred vs. unpreferred) is greater than the gap in any other property.

\subsection{Dynamic Degree analysis}
When evaluating dynamic degree on VBench, we divide the prompt test set of dynamic degree into static and dynamic. We use the same instruction of Listing~\ref{lst:llm_dynamic_instruction} to label dynamic and static propmts. We mark score 3 as dynamic and score 1, 2 as static. Then, we redefine dynamic degree by adding the portion of dynamic videos in filtered dynamic prompts and static videos in filtered static prompts.  
\subsection{User study details}
\label{subsec:userstudy_details}
Assessing the quality of generated content is often complicated by its inherent subjectivity. To support our findings and gain deeper insights into human preferences, we conducted a comprehensive user study involving 30 participants. Participants were given a prompt, and a set of videos, consisting of outputs from \ours, SFT-on, and Full. The participants were given 18 sets of the data from VideoCrafter 2 and 18 sets from AnimateDiff, resulting in a total of 1080 responses. They were asked to rank the overall preference of the videos based on three given criteria: 1-Visual Quality, 2-Motion Quality, and 3-Text Alignment. The samples used for the user study were chosen
randomly from a large, unbiased pool. An example question of the user study is provided in Figure~\ref{fig:userstudy_example}.

\section{Additional Experiments}
\label{sec:sup_additional_exp}
In this section, we present the results of additional experiments that could not be included in the main paper due to page constraints.

\subsection{SFT Weight Experiment}
\label{subsec:sftweight}
Since SFT Weight is a new parameter that we introduce, we conduct extensive experiments on VideoCrafter 2 in order to understand the effect of the parameter. We search through a total of 6 values, ranging from 1e2 to 1e7. The setup is identical to the main experiment, the only difference lying in $w_{SFT}$. 

\begin{table}[h]
    \centering
    \scalebox{0.85}{
    \begin{tabular}{lc|cccc}
        \toprule
        \textbf{Stage} & \textbf{weight} & \makecell{\textbf{Visual} \\ \textbf{Quality}} & \makecell{\textbf{Temporal} \\ \textbf{Consist.}} & \makecell{\textbf{Dynamic} \\ \textbf{Degree}} & \makecell{\textbf{Text} \\ \textbf{Align.}} \\
        \midrule
        \multirow{6}{*}{Stage 1} & 1e2 & 2.561 & 2.494 & 2.793 & 2.477 \\
        & 1e3 & 2.607 & 2.569 & 2.751 & 2.500 \\
        & 1e4 & 2.613 & 2.591 & 2.750 & 2.505 \\
        & 1e5 & 2.620 & 2.597 & 2.734 & 2.504 \\
        & 1e6 & \cellcolor{cyan!20}2.630 & \cellcolor{cyan!20}2.608 & \cellcolor{cyan!20}2.731 & \cellcolor{cyan!20}2.510 \\
        & 1e7 & 2.622 & 2.599 & 2.735 & 2.499 \\
        \midrule
        \multirow{6}{*}{Stage 2} & 1e2 & 2.392 & 2.292 & 2.775 & 1.982 \\
        & 1e3 & 2.424 & 2.349 & 2.763 & 2.016 \\
        & 1e4 & 2.621 & 2.601 & 2.737 & 2.518 \\
        & 1e5 & 2.634 & 2.618 & 2.720 & 2.516 \\
        & 1e6 & \cellcolor{cyan!20}2.629 & \cellcolor{cyan!20}2.617 & \cellcolor{cyan!20}2.728 & \cellcolor{cyan!20}2.518 \\
        & 1e7 & 2.449 & 2.403 & 2.712 & 1.959 \\
        \bottomrule
    \end{tabular}}
    \caption{\textbf{Ablation on SFT weight for VideoCrafter2.} The colored rows are the actual parameters used in the main experiment.}
    \label{tab:sftweight_vc}
\end{table}

Results in table~\ref{tab:sftweight_vc} demonstrate that SFT weight plays a crucial role in the performance of the models. While a typically low $w_{SFT}$ results in abnormally high dynamics that lead to video quality degradation, over a certain critical point, the model's overall performance just drops. Results show that such performance drop is more extreme in the second stage, meaning that the performance drop is likely resulted by overly fitting to the teacher's output as discussed in Section~\ref{subsec:motivation}. 

\begin{table}[h]
    \centering
    \scalebox{0.89}{
    \begin{tabular}{lccccc}
        \toprule
        \textbf{Stage} & \textbf{Method} & \makecell{\textbf{Visual} \\ \textbf{Quality}} & \makecell{\textbf{Temporal} \\ \textbf{Consist.}} & \makecell{\textbf{Dynamic} \\ \textbf{Degree}} & \makecell{\textbf{Text} \\ \textbf{Align.}} \\
        \midrule
        \multirow{2}{*}{1} & Pruned & 2.609 & 2.588 & 2.744 & 2.487 \\
        & \ours & 2.630 & 2.608 & 2.731 & 2.510 \\
        \midrule
        \multirow{2}{*}{2} & Pruned & 2.627 & 2.595 & 2.725 & 2.486 \\ 
        & \ours & 2.629 & 2.617 & 2.728 & 2.518 \\
        \midrule
        \multirow{2}{*}{3} & Pruned & 2.436 & 2.372 & 2.749 & 1.923 \\  
        & \ours & 2.594 & 2.580 & 2.718 & 2.429 \\
        \midrule
        & Full & 2.627 & 2.602 & 2.728 & 2.491 \\
        \bottomrule
    \end{tabular}}
    \caption{\textbf{Experimental results on VideoCrafter 2.} While our method shows consistent performance in recovering the degraded performance due to pruning, when VC2 is pruned for a third stage, the performance drops drastically, making it unreasonable to report the numbers. However, even so, \ours show great recovery of performance.}
    \label{tab:vc_stage3}
\end{table}

\subsection{Further Experiments for VideoCrafter2}
In this section, we present the results of training VC up to Stage 3.
As shown in Table~\ref{tab:vc_stage3}, after pruning two additional U-Net blocks, the performance of the pruned model drops significantly. Despite further training, the model fails to reach optimal performance. However, as observed in the table, after applying ReDPO, the performance gap dramatically improves, demonstrating that even with severe performance degradation, ReDPO and VIP effectively facilitate learning and recovery.

\subsection{Experiments for step-distilled model}
\label{subsec:step_distill}
In this section, to demonstrate \ours's effectiveness on a step-distilled model, we experiment on AnimateDiff Lightning\cite{lin2024animatediff_lightning}, a 4-step distilled model. As shown in Table~\ref{tab:videoreward}, our method meets the performance of the full model in both stages, which is remarkable considering that it has already been distilled once. Contrarily, Table~\ref{tab:adlightning} show that SFT struggles significantly, even with the same \ours framework with only a difference in loss. These findings underscore the robustness of \ours, especially in heavily pruned, capacity-limited settings like step-distilled models. The clear advantage over SFT in such scenarios emphasizes the effectiveness of our targeted, preference-driven distillation strategy.

\begin{table}[h]
    \centering
    \scalebox{0.78}{
    \begin{tabular}{lc|cccc|c}
        \toprule
        \textbf{Stage} & \textbf{Method} & 
        \makecell{\textbf{Visual} \\ \textbf{Quality}} & \makecell{\textbf{Temporal} \\ \textbf{Consist.}} & \makecell{\textbf{Dynamic} \\ \textbf{Degree}} & \makecell{\textbf{Text} \\ \textbf{Align.}} &
        \textbf{AVG.} \\
        \midrule
         & Full & \textbf{2.644} & 2.560 & 2.542 & \textbf{2.411} & 2.539 \\
        \cmidrule{1-7}
        \multirow{2}{*}{S1} & Pruned & 2.640 & 2.566 & 2.532 & 2.410 & 2.537 \\
        & \cellcolor{cyan!20}\loss & \cellcolor{cyan!20}2.642 & \cellcolor{cyan!20}2.562 & \cellcolor{cyan!20}2.537 & \cellcolor{cyan!20}2.410 & \cellcolor{cyan!20}2.538 \\
        \cmidrule{1-7}
        \multirow{2}{*}{S2} & Pruned & 2.640 & 2.564 & 2.535 & 2.393 & 2.533 \\
        & \cellcolor{cyan!20}\loss & \cellcolor{cyan!20}2.641 & \cellcolor{cyan!20}\textbf{2.565} & \cellcolor{cyan!20}\textbf{2.550} &  \cellcolor{cyan!20}2.397 & \cellcolor{cyan!20}2.538 \\
        \bottomrule
    \end{tabular}}
    \caption{Experiments on AD Lightning with VideoReward.}
    \label{tab:videoreward}
\end{table}

\subsection{Experiments for reward model}
\label{subsec:reward_model}
In this section, to examine the robustness of our method across different reward models, we replaced VideoScore with a more recent reward model, VideoReward\cite{liu2025improving}, in our two-stage pruning experiment on AnimateDiff Lightning. As shown in Table~\ref{tab:adlightning}, using VideoReward led to improved performance compared to VideoScore. This result highlights that our framework is \textit{reward-model agnostic}-it uses reward models only to generate preference pairs, without any direct propagation of reward values. Consequently,improvements in the reward models translate directly into better distillation outcomes.

\begin{table}[h]
    \centering
    \scalebox{0.8}{
    \begin{tabular}{l|cccc}
        \toprule
        \textbf{Loss} & \makecell{\textbf{Visual} \\ \textbf{Quality}} & \makecell{\textbf{Temporal} \\ \textbf{Consist.}} & \makecell{\textbf{Dynamic} \\ \textbf{Degree}} & \makecell{\textbf{Text} \\ \textbf{Align.}} \\
        \midrule
        SFT & 2.637 & \textbf{2.572} & 2.525 & 2.388 \\
        \rowcolor{cyan!20}\textbf{\loss}(videoscore) & \textbf{2.641} & 2.564 & \underline{2.540} & \underline{2.396} \\
        \rowcolor{cyan!20}\textbf{\loss}(videoreward) & \textbf{2.641} & \underline{2.565} & \textbf{2.550} & \textbf{2.397} \\ 
        \bottomrule
    \end{tabular}}
    \caption{Experiments on AnimateDiff Lightning.}    
    \label{tab:adlightning}
\end{table}

\section{Additional Explanations on Motivation}
\label{sec:additional_motivation}

\begin{figure}[th]
    \centerline{\includegraphics[width=0.48\textwidth]{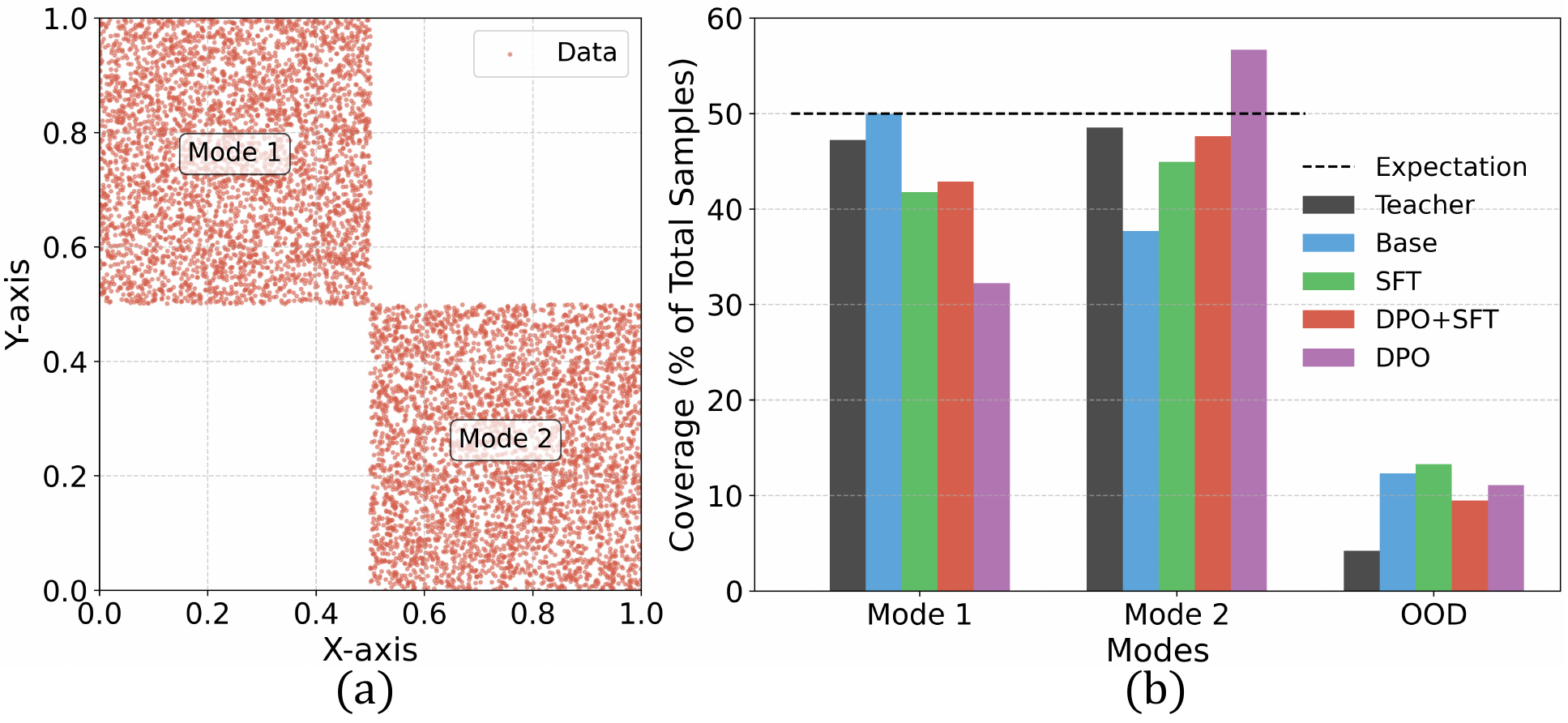}}
    \caption{\textbf{Analysis of toy experiment results.} (a) Ground-truth distribution used in the toy experiment. (b) Number of samples assigned to each mode and the number of out-of-distribution (OOD) samples.} 
    \label{fig:supp_motivation} 
\end{figure}

In this section, we illustrate the limitations of conventional knowledge distillation methods in diffusion models, which rely on SFT loss—particularly when applied to capacity-constrained student models. We then propose an alternative distillation approach and validate its effectiveness through a controlled toy experiment.

Conventional knowledge distillation methods for diffusion models typically transfer knowledge from the teacher to the student by directly minimizing SFT loss. However, in models with limited capacity, this objective forces the student to align with the teacher as closely as possible, often resulting in distributional averaging and overly smoothed outputs. This occurs because minimizing the SFT loss implicitly prioritizes fitting the mean over preserving sharpness~\cite{mse_bad_perception_distortion, mse_bad_3_wasserstein}. To address this, it is crucial to provide explicit guidance—here, using DPO~\cite{rafailov2023direct}—that prioritizes important features, ensuring the student allocates its limited capacity effectively rather than blindly mimicking the teacher.

To investigate this, we conducted a toy experiment by training a high-capacity \textit{teacher} and a low-capacity \textit{base student} on a two-dimensional dataset. We implemented the teacher model’s diffusion backbone as a 4-layer MLP with a hidden dimension of 64. Since pruning small MLPs does not behave similarly to pruning large U-Nets—where the goal is typically initializing the student model to closely resemble the teacher—we trained a separate, smaller-scale student model, named the base student, to replicate this phenomenon. Specifically, we trained the base student with a diffusion backbone consisting of a 2-layer MLP with a hidden dimension of 32.

Both models learn to approximate the data distribution shown in Figure~\ref{fig:supp_motivation}~(a). However, the base student exhibits an imbalanced learned distribution—as illustrated in Figure~\ref{fig:supp_motivation}~(b)—due to its limited capacity. We distilled the teacher’s knowledge into the base student using three distinct loss variants: $L_{SFT}$, $L_{DPO}$, and $L_{DPO} + L_{SFT}$. The $L_{SFT}$ loss minimizes the $L_2$ distance between predictions, while $L_{DPO}$ explicitly prioritizes Mode 2 to address the base student’s difficulty in generating samples in this region.

To construct the $L_{DPO}$-based variants, we first created a DPO dataset by setting the reward function to prioritize samples in Mode 2. A preference dataset was then built, selecting winning samples from the teacher within Mode 2 and losing samples from the student outside Mode 2. Using this dataset, we trained both the $L_{DPO}$ student and the $L_{DPO} + L_{SFT}$ student. The only difference among all trained models was their loss formulation.

\begin{figure}[t]
    \centerline{\includegraphics[width=0.48\textwidth]{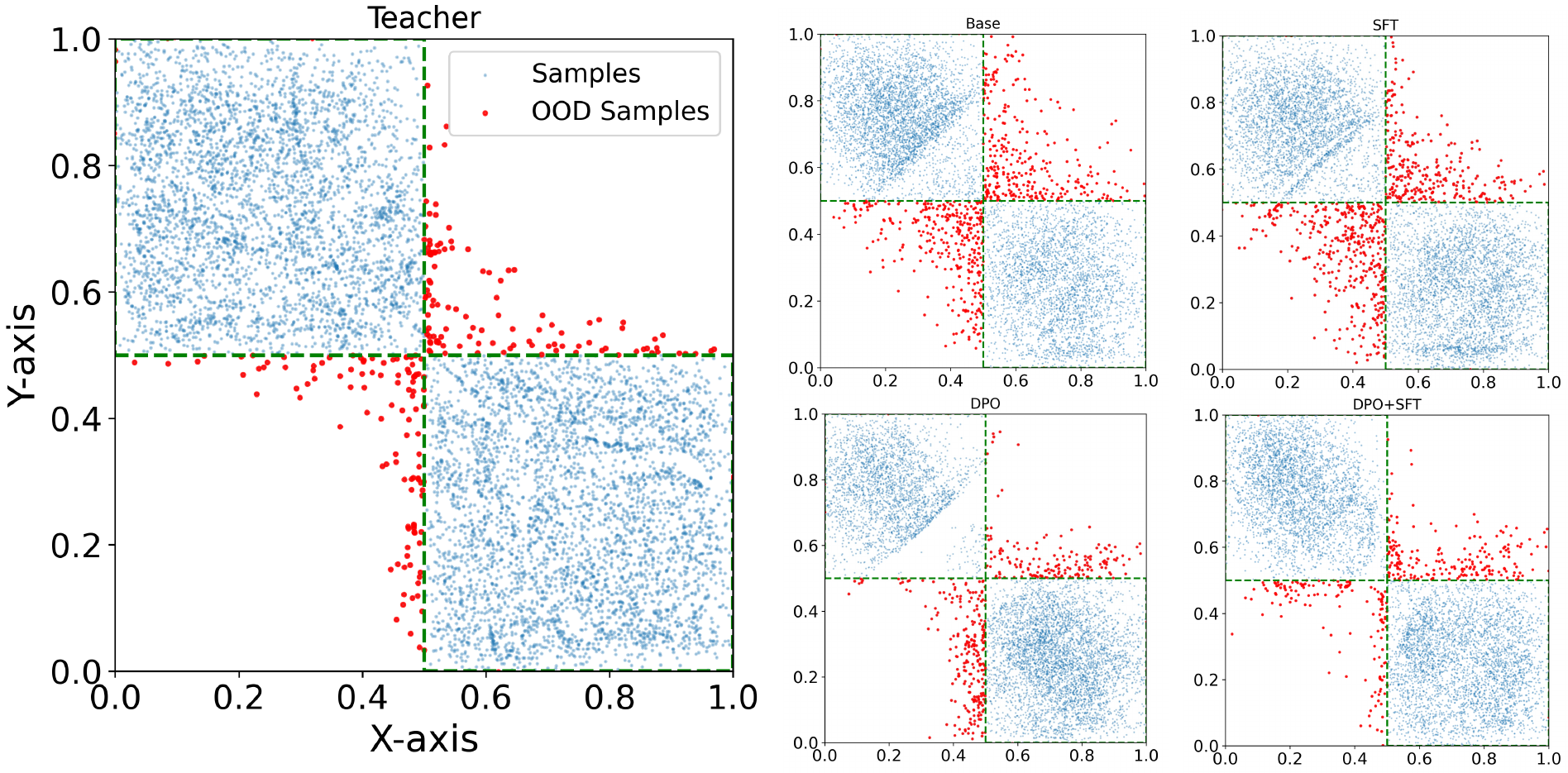}}
    \caption{\textbf{Visualization of learned distributions.} Combining DPO with SFT yields a distribution that most closely aligns with the teacher distribution.}
    \label{fig:supp_motivation_2} 
\end{figure}

Figure~\ref{fig:supp_motivation} and Figure~\ref{fig:supp_motivation_2} presents three key findings: (1) Distillation using $L_{SFT}$ leads to excessive distributional smoothing, resulting in more out-of-distribution (OOD) samples than even the base student. (2) Distillation using $L_{DPO}$ reallocates model capacity toward favoring Mode 2; however, due to inherent over-optimization issues, the model becomes misdirected in uncertain regions of the reward distribution—specifically demonstrated by degradation in Mode 1. (3) Combining $L_{DPO}$ with $L_{SFT}$ balances these effects, effectively prioritizing Mode 2 while mitigating over-optimization. As a result, it reduces the number of OOD samples compared to the other methods and more closely follows the teacher distribution. These observations suggest that $L_{DPO}$ facilitates efficient reallocation of the model’s limited capacity toward critical generative properties, while avoiding over-optimization. 

\begin{figure}[b]
    \centerline{\includegraphics[width=0.7\linewidth]{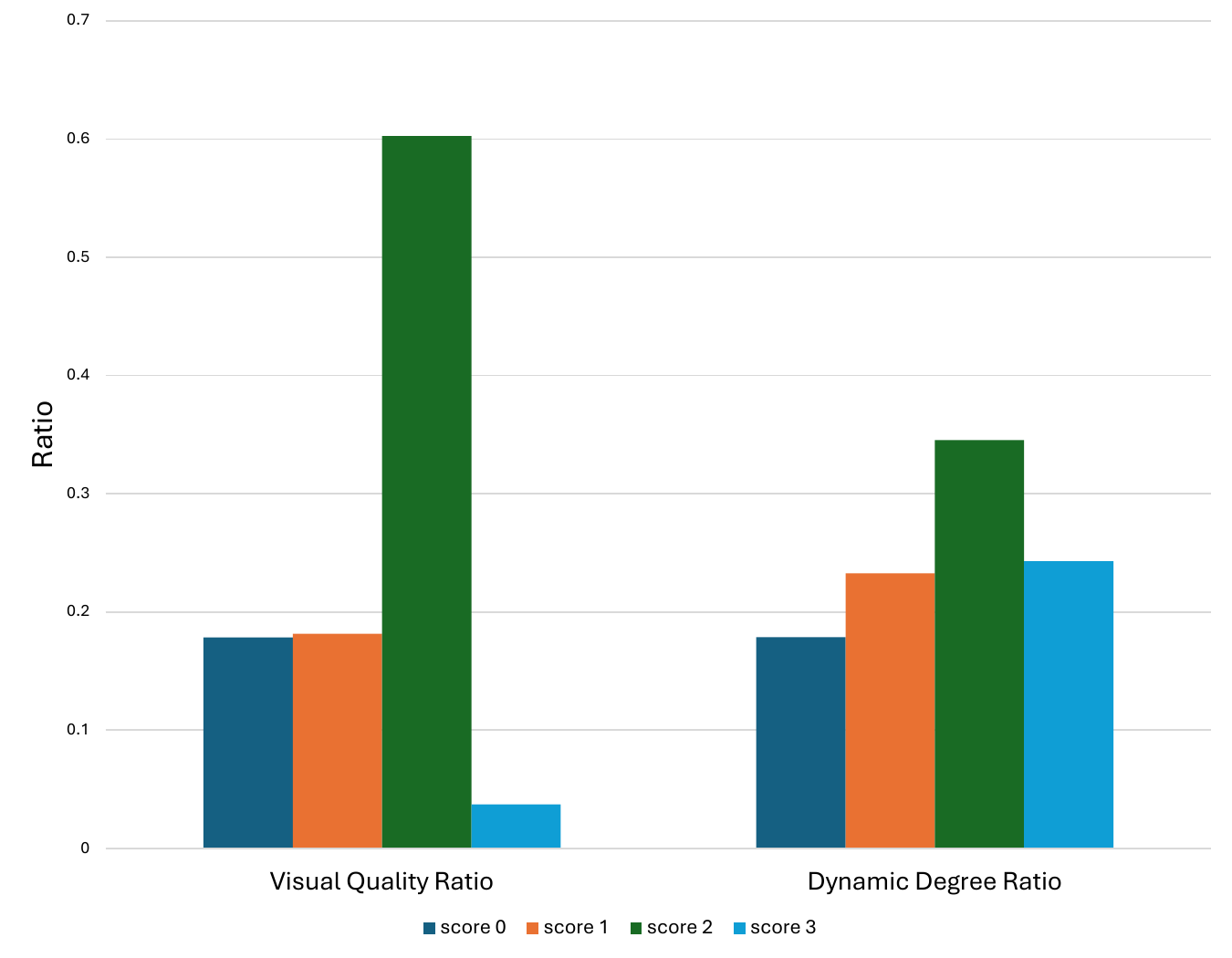}}
    \caption{The score distribution of Dynamic Degree and Visual Quality.}
    \label{fig:score_distribution} 
\end{figure}

\begin{figure}[b]
    \centering
    \includegraphics[width=1.0\linewidth]{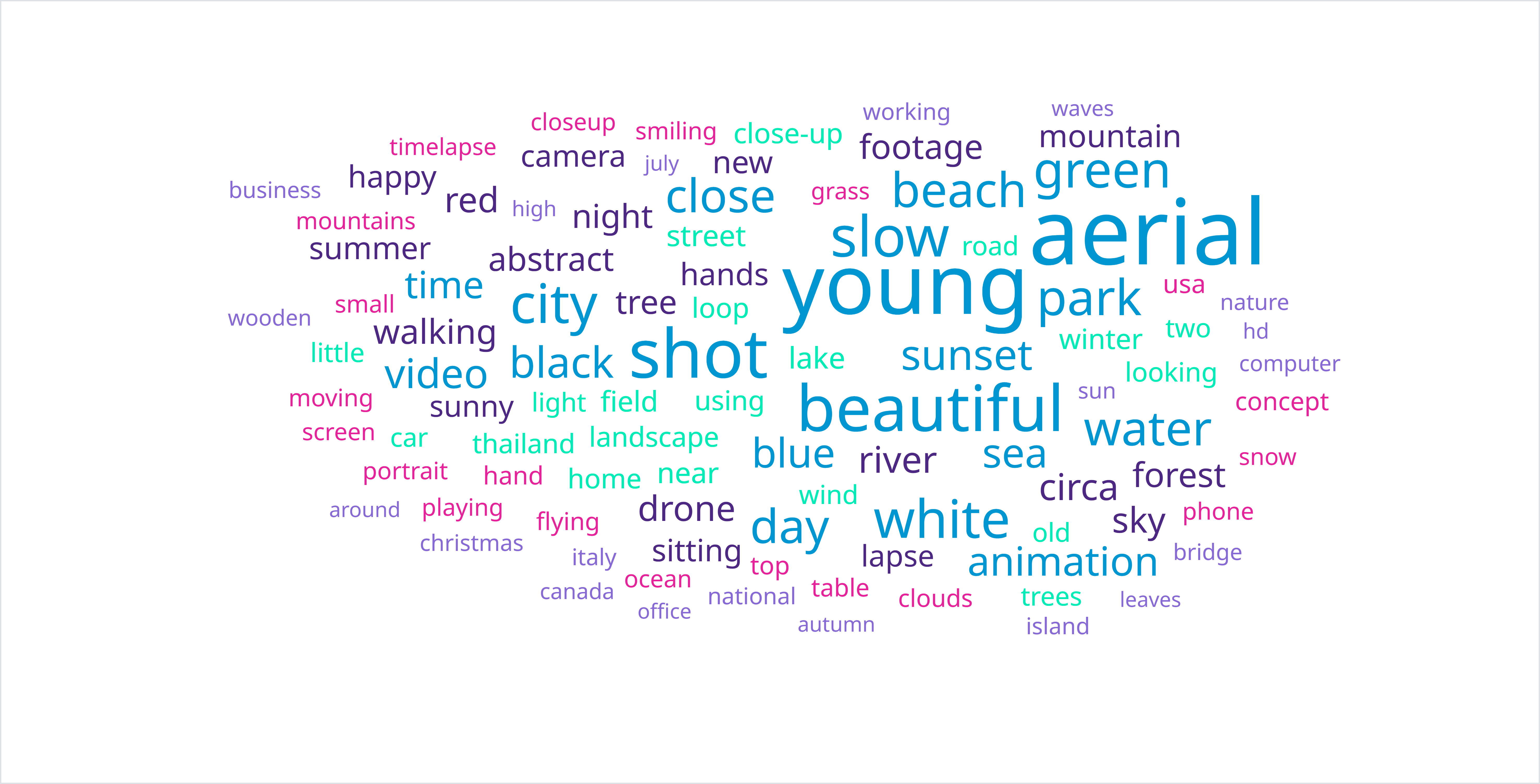}\\
    \includegraphics[width=1.0\linewidth]{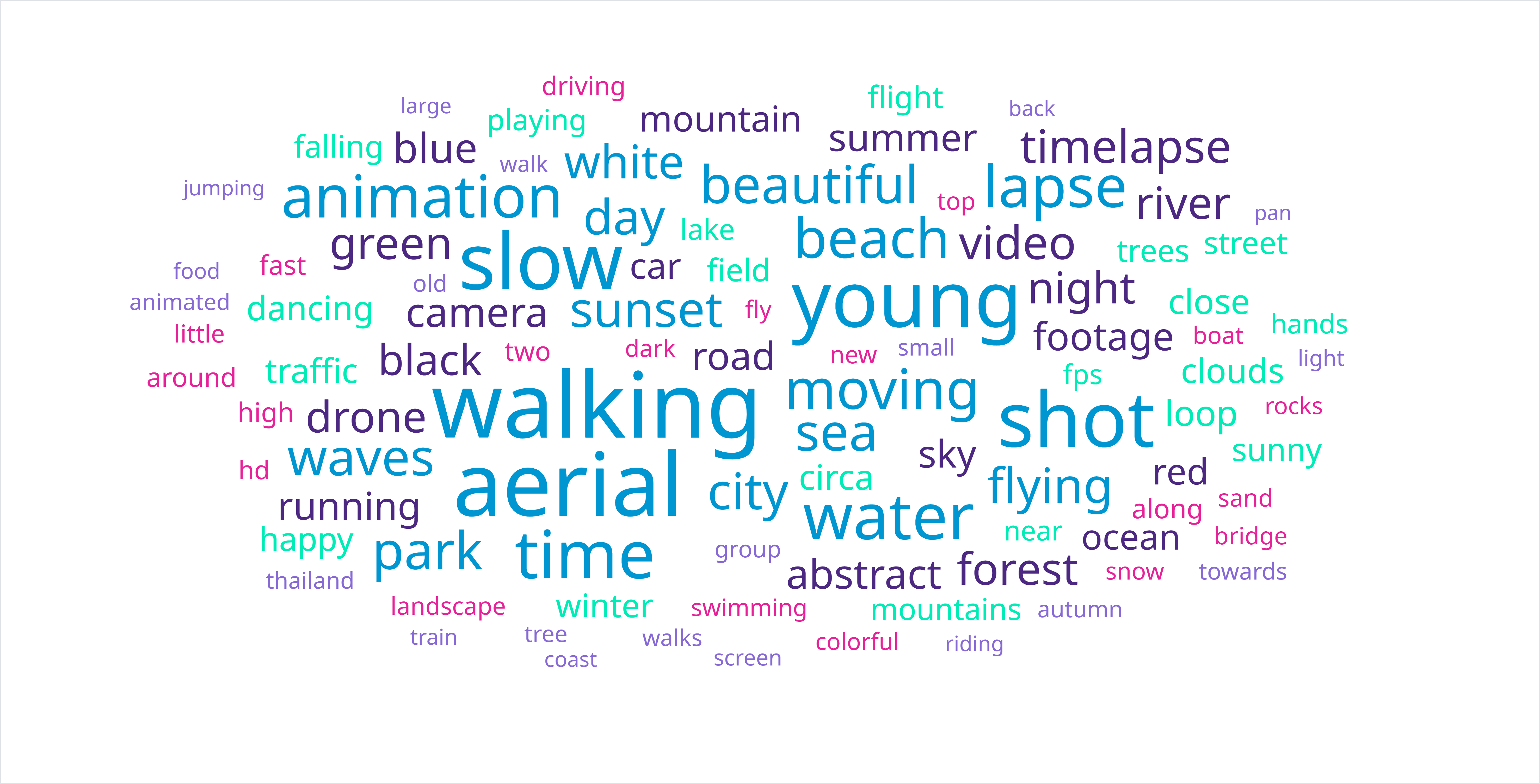}\\
    \includegraphics[width=1.0\linewidth]{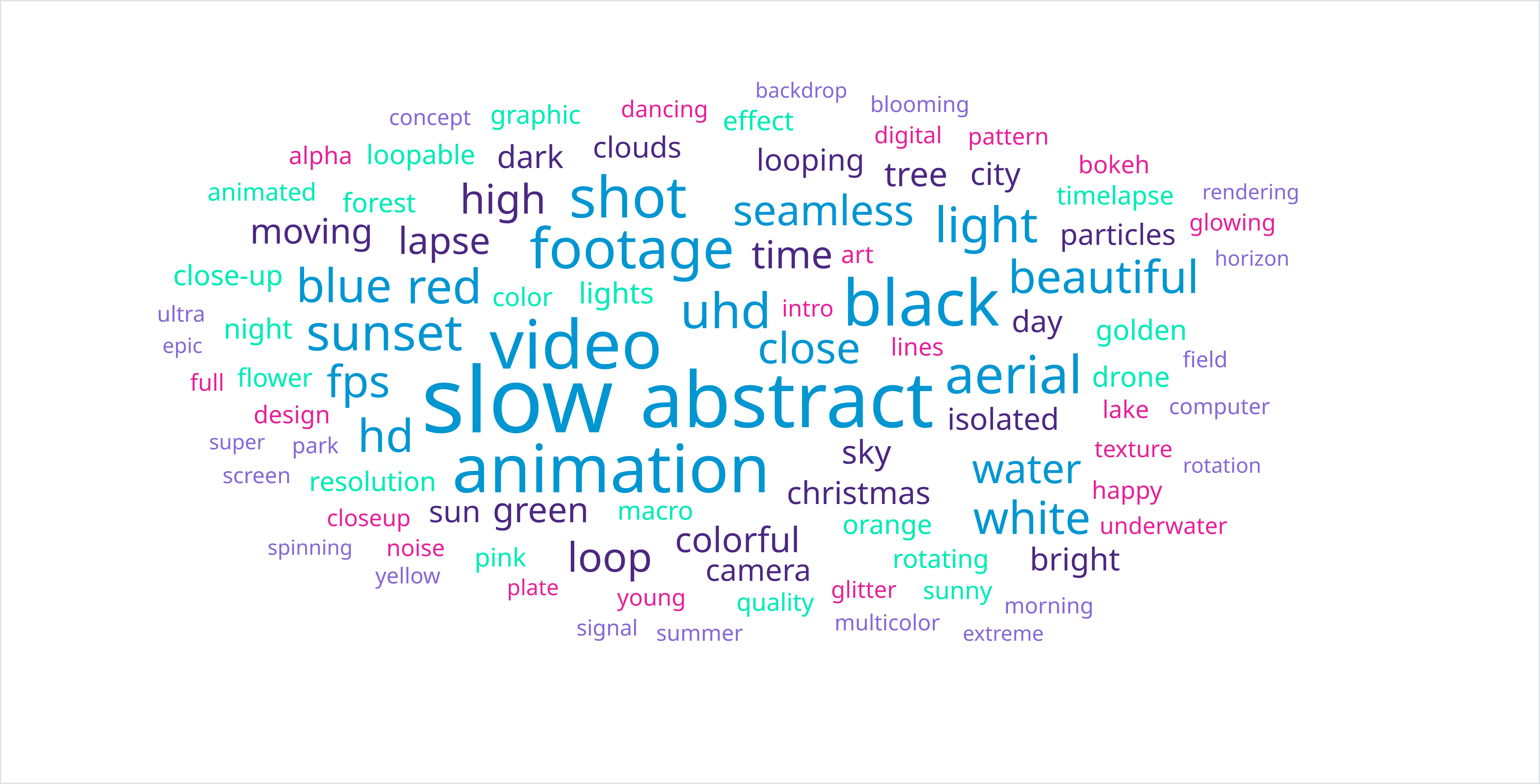}
    \caption{\textbf{Word clouds of prompt sets.} The word cloud of prompt set before LLM filtering (Top). After filtering, the word cloud of dynamic quality (Middle) and visual quality (Bottom) grounded to its property each.}
    \label{fig:word_cloud}
\end{figure}

\section{Prompt filtering}
\label{sec:prompt_filter}
For prompt filtering, we use Gemini 2.0 Flash to score each prompt from 0 to 3. The score distribution of two properties after LLM filtering is shown in Figure~\ref{fig:score_distribution}. The word cloud of prompts before and after filtering is shown in Figure~\ref{fig:word_cloud}.

\subsection{Dynamic Degree}
As shown in Listing~\ref{lst:llm_dynamic_instruction}, we designed an LLM-based filtering process to assign dynamic motion scores to prompts. We instruct LLM to score 1 if the prompt contains no motion, score 2 if the prompt contains minimal motion, and score 3 if the prompt contains considerable motion. When configuring an example of dynamic degree on an instruction, we use an evaluation set of prompt from VBench. It contains multiple prompts with dynamic motions. We select the score 3 prompt in Dynamic Degree for video curation. 

\subsection{Visual Quality}
As shown in Listing~\ref{lst:llm_visual_instruction}, we designed an LLM-based filtering process to assign visual quality scores to the prompts. We instruct LLM to score 1 if the prompt contains simple or generic descriptions, score 2 if the prompt contains moderate visual attributes, and score 3 if the prompt contains highly descriptive, rich in visual attributes. When configuring an example of visual quality, we use LLM to generate appropriate examples. We select the score 3 prompt in Visual Quality for video curation. %

\subsection{Text Alignment}
To filter a prompt set that enhances text alignment, we hypothesize that high-quality text prompts are essential for generating videos that accurately capture semantic meaning. To ensure quality, we establish criteria to exclude prompts that are too short or too long, overly complex, or dominated by location names. Specifically, we retain single-sentence prompts with 5 to 25 words, excluding articles. Additionally, we employ LLM-based filtering during the initial selection stage by assigning a score of 0 to eliminate unusable prompts. By applying these constraints, we ensure that the input prompt set maintains linguistic clarity and relevance, facilitating the construction of a dataset optimized for text alignment. We select non-zero score prompts from Dynamic Degree and Visual Quality.

\section{Qualitative results}
\label{sec:supp_quan}
We additionally report qualitative results of our work from Figure~\ref{fig:vc4} to Figure~\ref{fig:ad8}.

\begin{table*}[h]
    \centering
    \begin{minipage}{\textwidth}
        \begin{lstlisting}[style=mystyle, caption={LLM Instruction for Dynamic Motion Scoring}, label={lst:llm_dynamic_instruction}]
###Task Overview:
You are a model responsible for scoring prompts for a video diffusion model. 
Your job is to evaluate and determine the level of dynamic motion present in each prompt.
The final output should be a score from 0 to 3.

### Task Description:
1. Assign score 0 if the prompt is unusable due to:
   - Fragmented, unclear, or incoherent sentences.
   - Excessive mentions of country names (distracts from motion evaluation).

2. Otherwise, analyze the degree of motion and assign a score from 1 to 3:
   - 1: Static Scene -> No motion or movement (e.g., a still scene, a stationary object).
   - 2: Minimal Motion -> Slight transitions or small repetitive actions (e.g., a person blinking, tree leaves rustling, a slow tilt upward).
   - 3: Considerable Motion -> Significant movement or scene transformation (e.g., running, a car driving, waves crashing, person walking, a smooth tracking shot following person).

### Examples:
- 1: A still painting of a landscape with a sunset.
- 2: A person slowly turning the pages of a book.
- 3: A cyclist racing through a city, dodging traffic.

### Output format:
Always return your result in this format:
[RESULT] <a score between 0 and 3>
        \end{lstlisting}
    \end{minipage}
\end{table*}

\begin{table*}[t]
    \centering
    \begin{minipage}{\textwidth}
        \begin{lstlisting}[style=mystyle, caption={LLM Instruction for Visual Quality Scoring}, label={lst:llm_visual_instruction}]
###Task Overview:
You are a model responsible for scoring prompts for a video diffusion model. 
The definition of "Visual Quality" is the quality of the video in terms of clearness, 
resolution, brightness, and color. Your job is to evaluate and determine the level of 
Visual Quality present in each prompt. The final output should be a score from 0 to 3.

### Task Description:
1. Assign score 0 if the prompt is unusable due to:
   - Fragmented, unclear, or incoherent sentences.
   - Excessive mentions of country names (distracts evaluation).

2. Otherwise, analyze the degree of Visual Quality and assign a score from 1 to 3:
   - Score 1: Low Visual Quality: Vague or generic descriptions with minimal details. No mention of visual attributes like lighting, colors, resolution, or atmosphere.
   - Score 2: Moderate Visual Quality: Some visual attributes are present but lack specificity or coherence. Colors, lighting, and resolution are mentioned but not in depth.
   - Score 3: High Visual Quality: The prompt is highly descriptive, rich in visual attributes. Specific details about lighting, resolution, colors, textures, and clarity are included.

### Examples:
- Score 1: A beach with waves.
- Score 2: A snow-covered mountain with a few clouds in the sky.
- Score 3: An elderly man sitting on a worn leather armchair beside a crackling fireplace, the warm glow casting deep shadows on the wooden walls.

### Output format:
Always return your result in this format:
[RESULT] <a score between 0 and 3>
        \end{lstlisting}
    \end{minipage}
\end{table*}

\begin{figure*}[!h]
    \centering
    \includegraphics[width=0.45\textwidth]{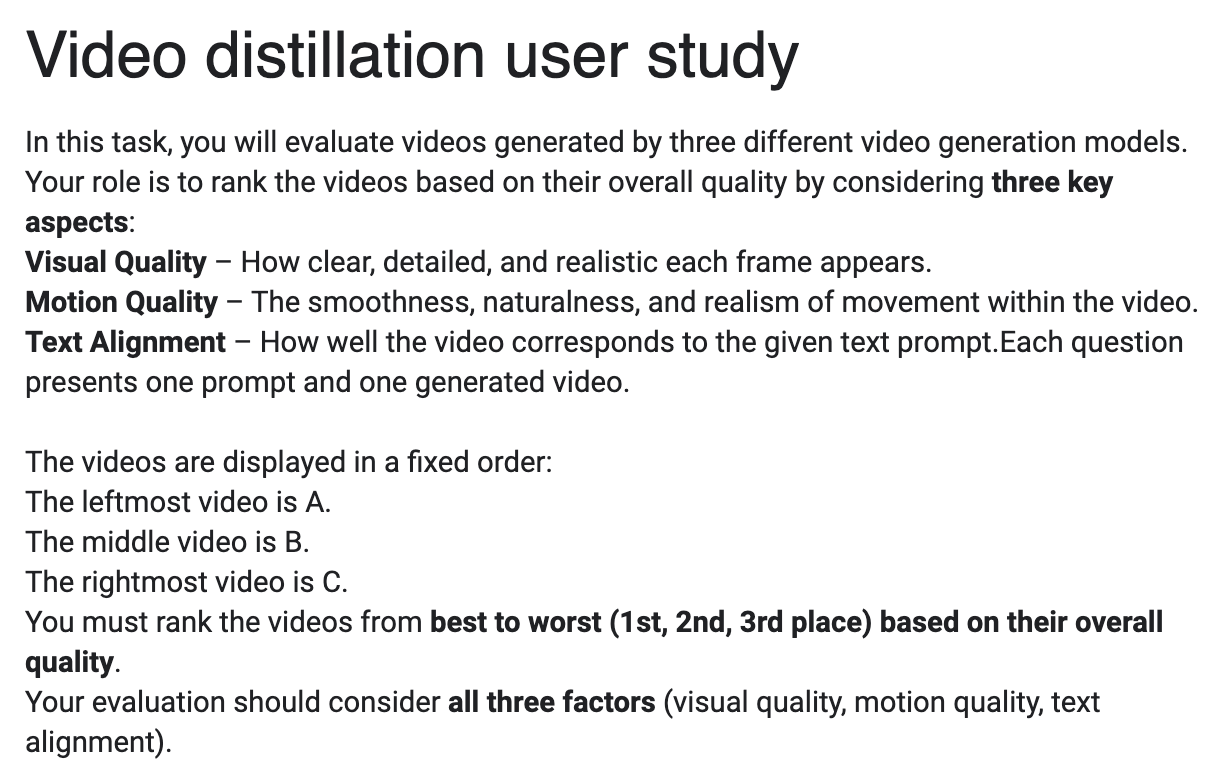}\hfill
    \includegraphics[width=0.45\textwidth]{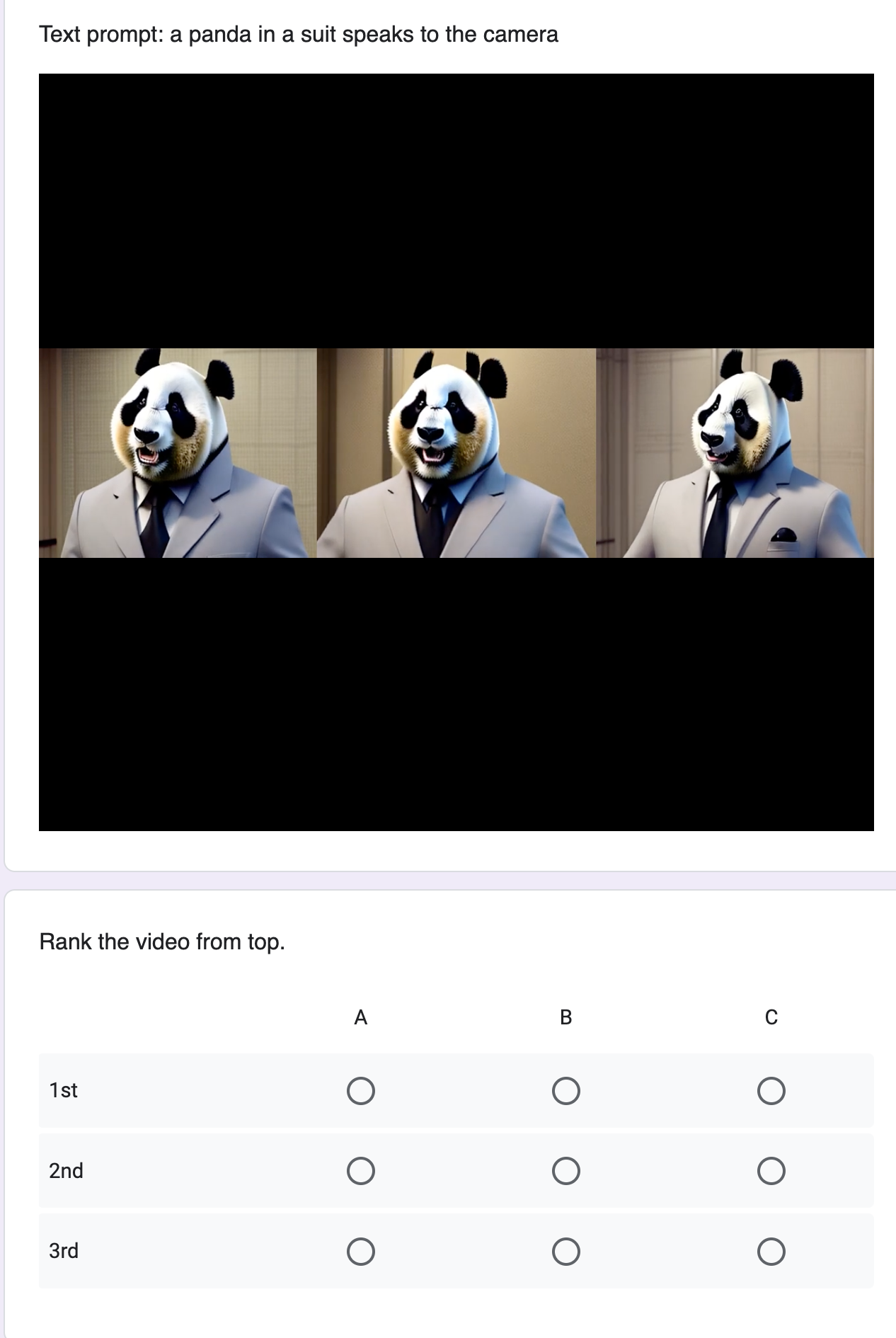}
    \caption{Example of the user study instructions that the participants received and a sample of an actual question.}
    \label{fig:userstudy_example}
\end{figure*}

\begin{figure*}[!h]
    \centering
    \includegraphics[width=0.8\textwidth]{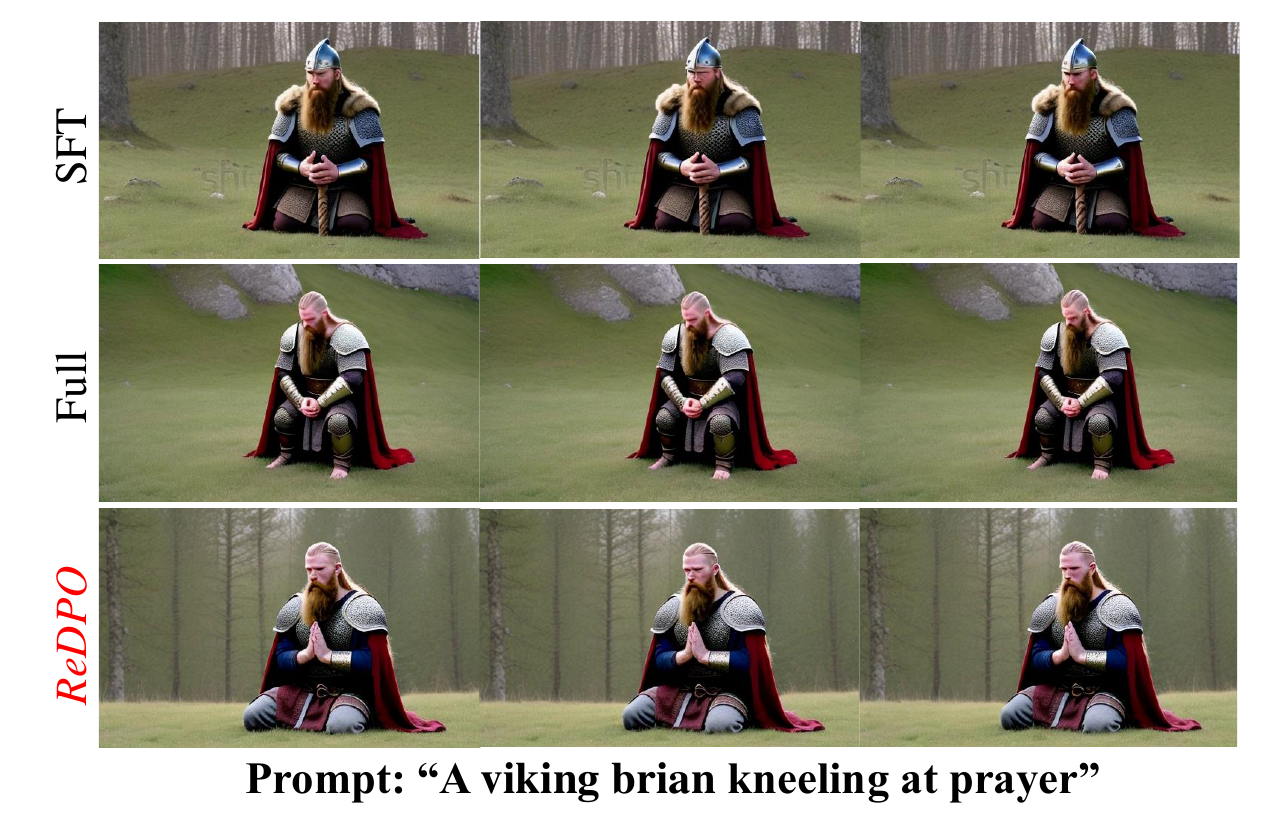}
    \caption{Qualitative example of VideoCrafter 2}
    \label{fig:vc4}
\end{figure*}

\begin{figure*}[!h]
    \centering
    \includegraphics[width=0.8\textwidth]{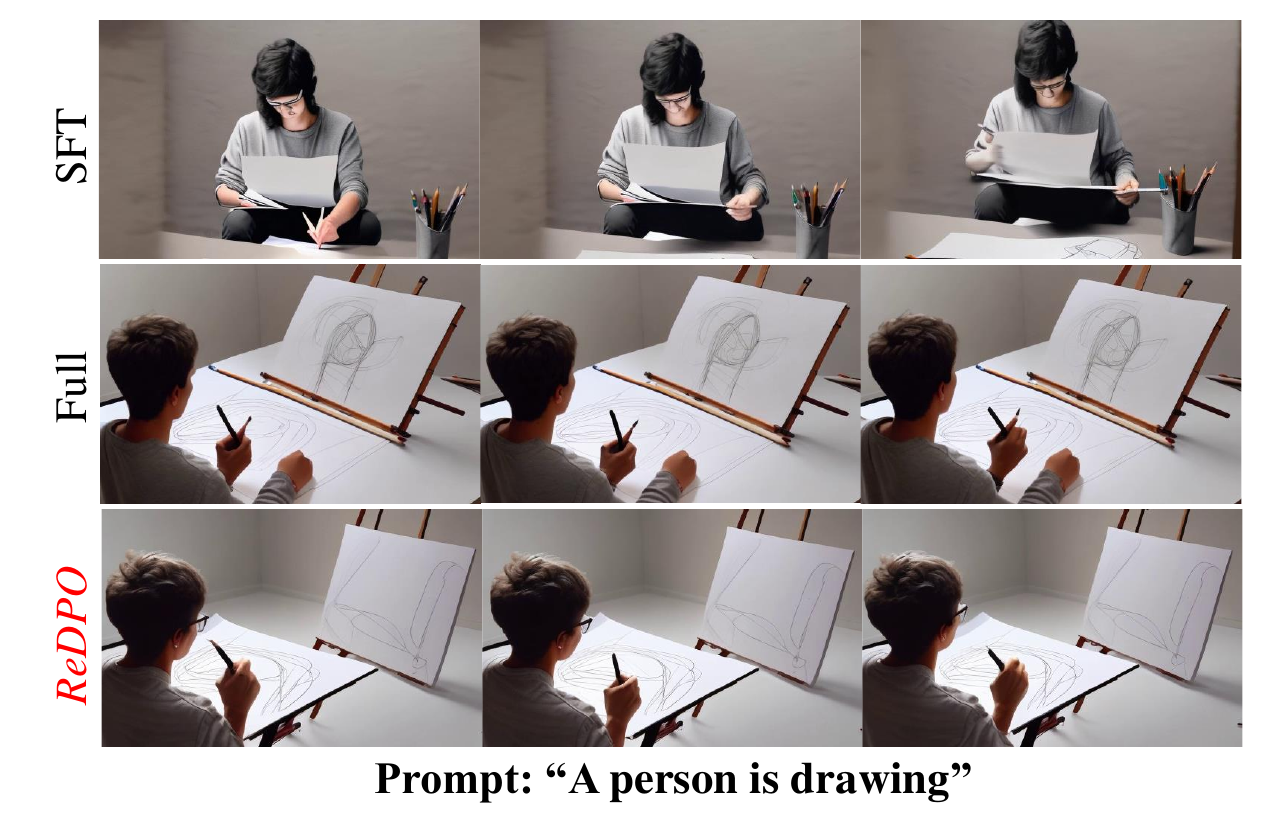}
    \caption{Qualitative example of VideoCrafter 2}
    \label{fig:vc3}
\end{figure*}

\begin{figure*}[!h]
    \centering
    \includegraphics[width=0.8\textwidth]{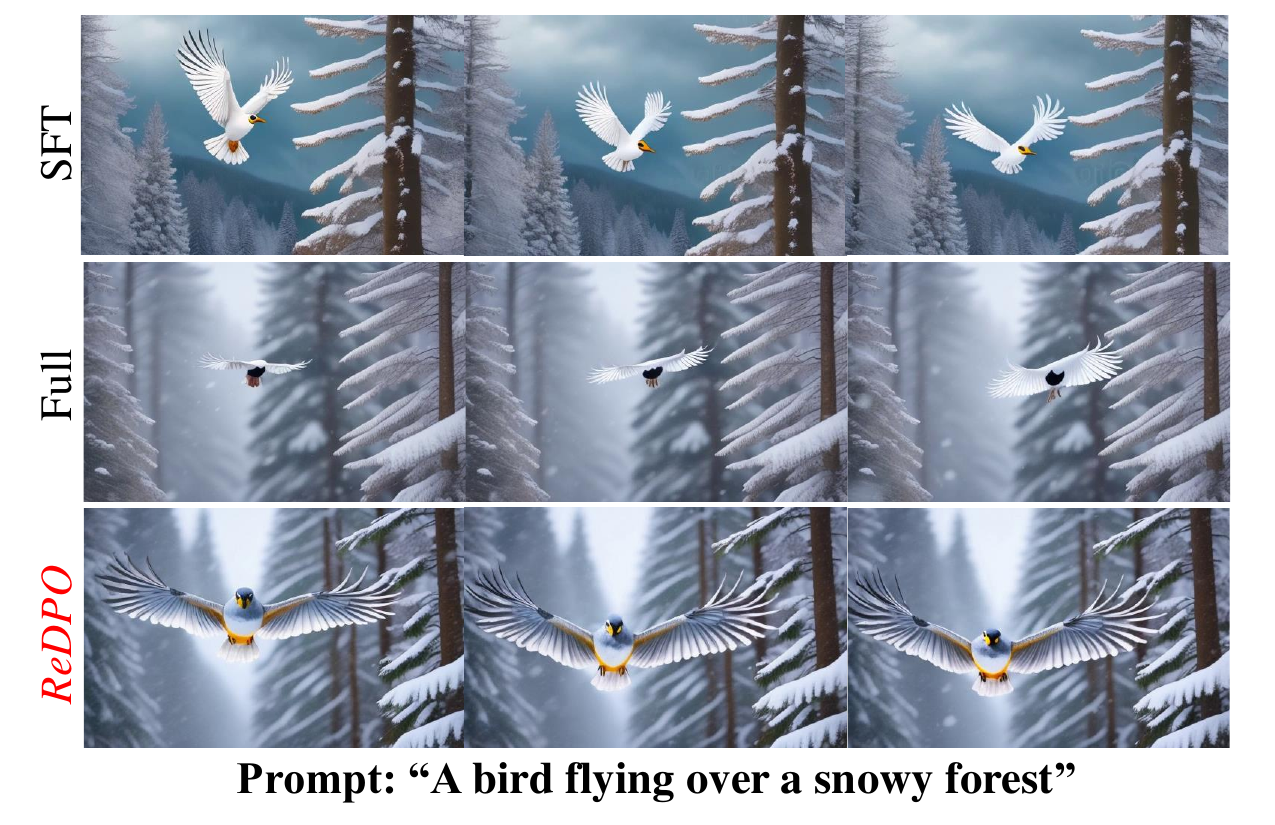}
    \caption{Qualitative example of VideoCrafter 2}
    \label{fig:vc1}
\end{figure*}

\begin{figure*}[!h]
    \centering
    \includegraphics[width=0.8\textwidth]{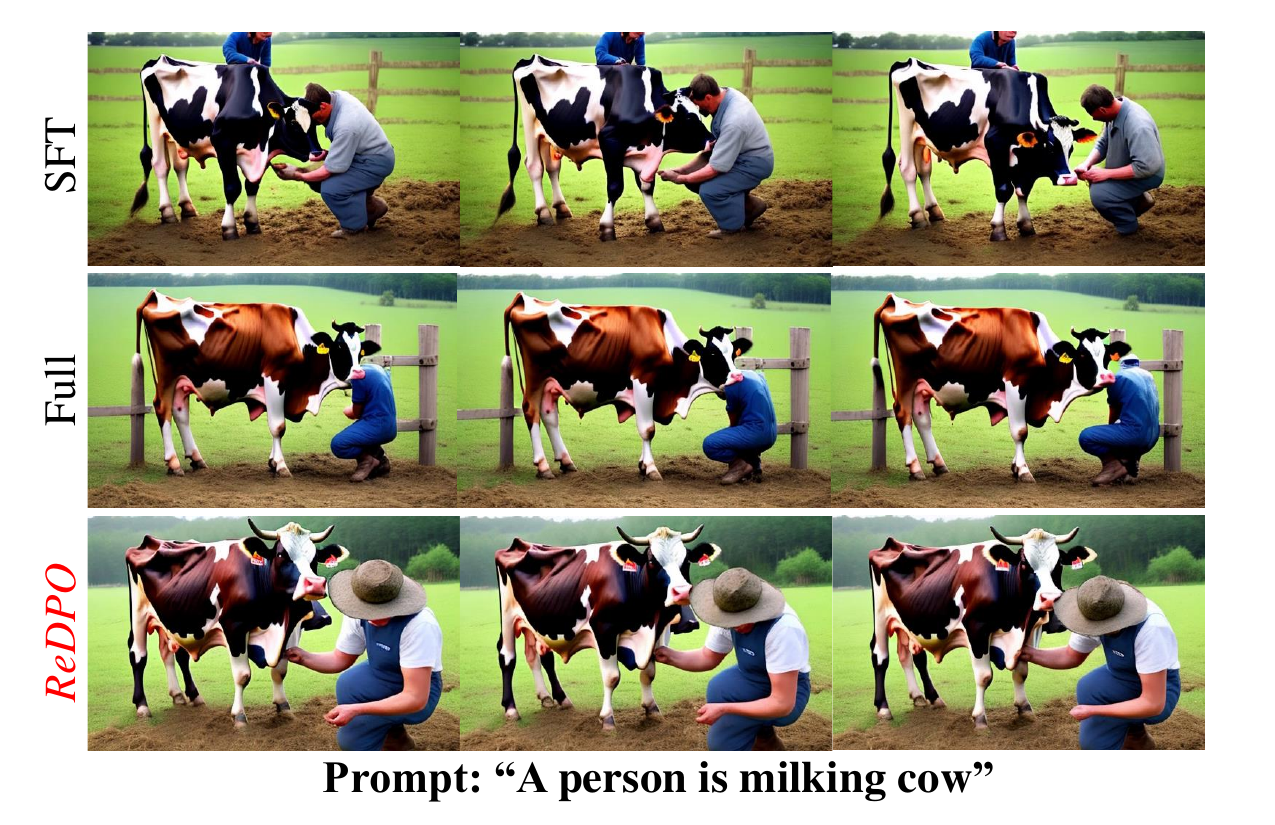}
    \caption{Qualitative example of VideoCrafter 2}
    \label{fig:vc2}
\end{figure*}

\begin{figure*}[!h]
    \centering
    \includegraphics[width=0.8\textwidth]{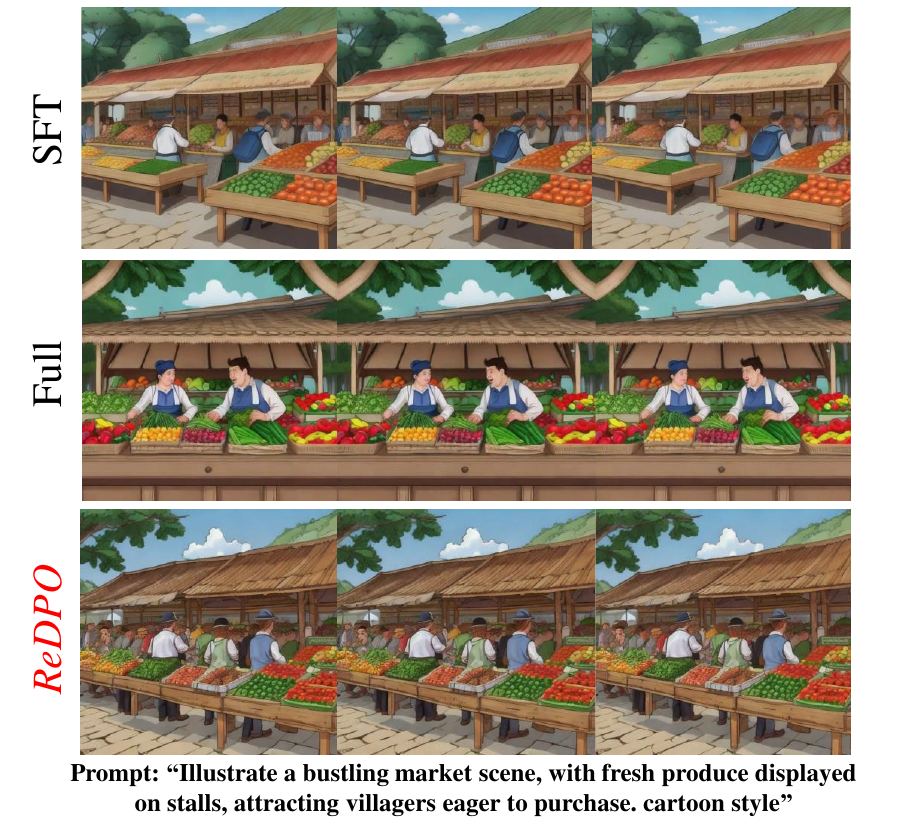}
    \caption{Qualitative example of AnimateDiff}
    \label{fig:ad1}
\end{figure*}

\begin{figure*}[!h]
    \centering
    \includegraphics[width=0.8\textwidth]{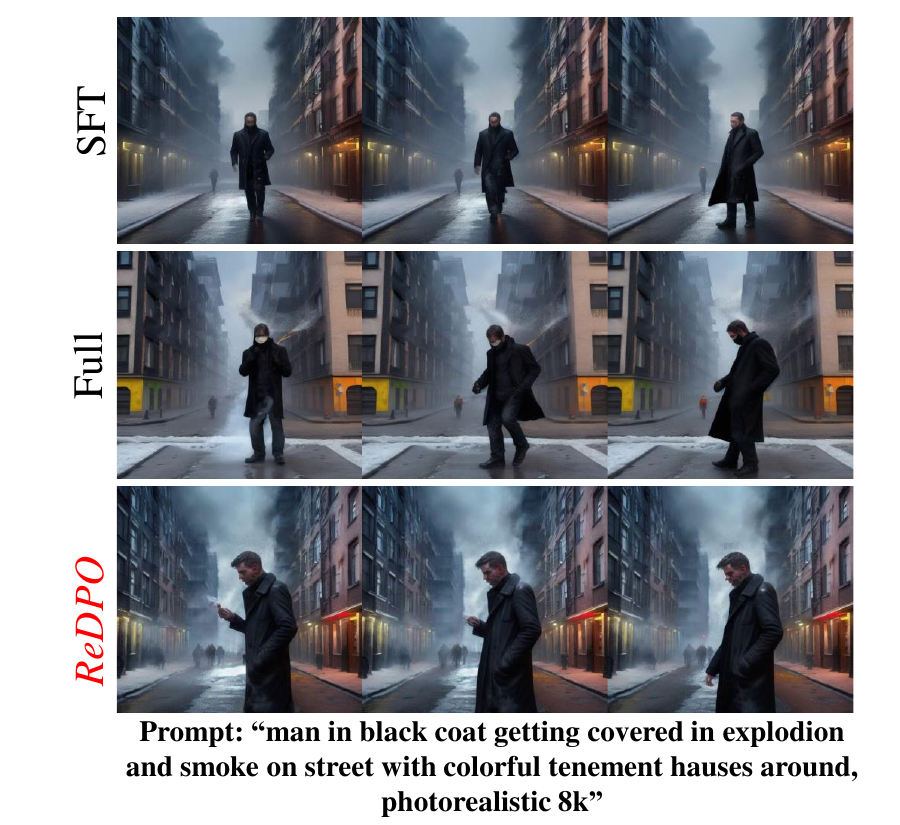}
    \caption{Qualitative example of AnimateDiff}
    \label{fig:ad2}
\end{figure*}

\begin{figure*}[!h]
    \centering
    \includegraphics[width=0.8\textwidth]{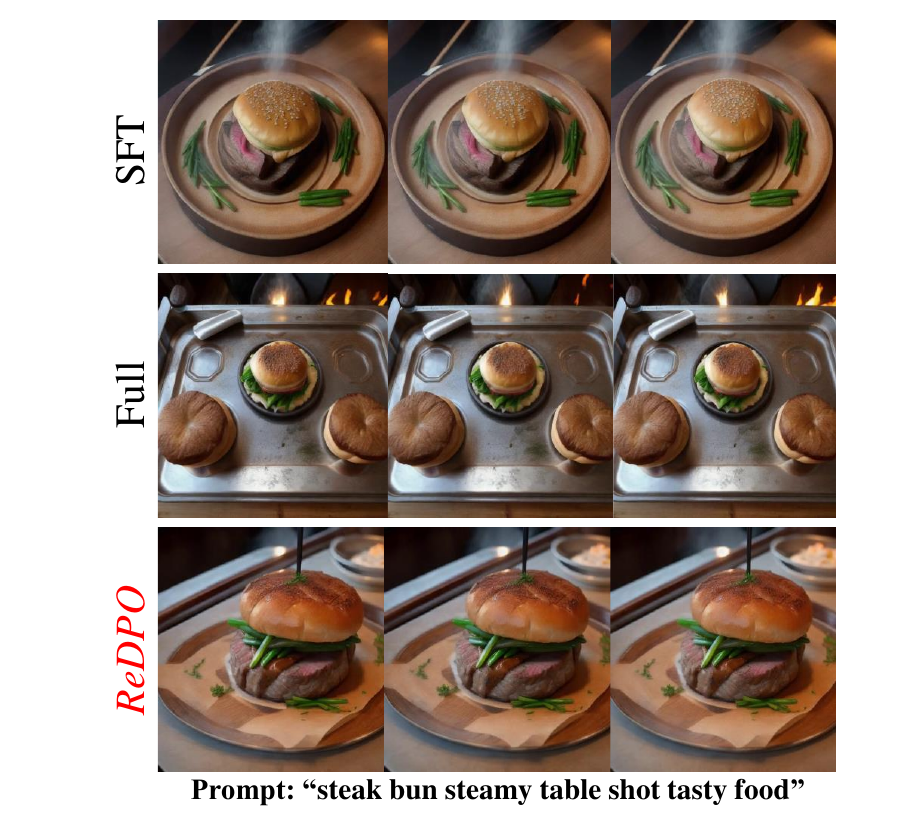}
    \caption{Qualitative example of AnimateDiff}
    \label{fig:ad4}
\end{figure*}

\begin{figure*}[!h]
    \centering
    \includegraphics[width=0.8\textwidth]{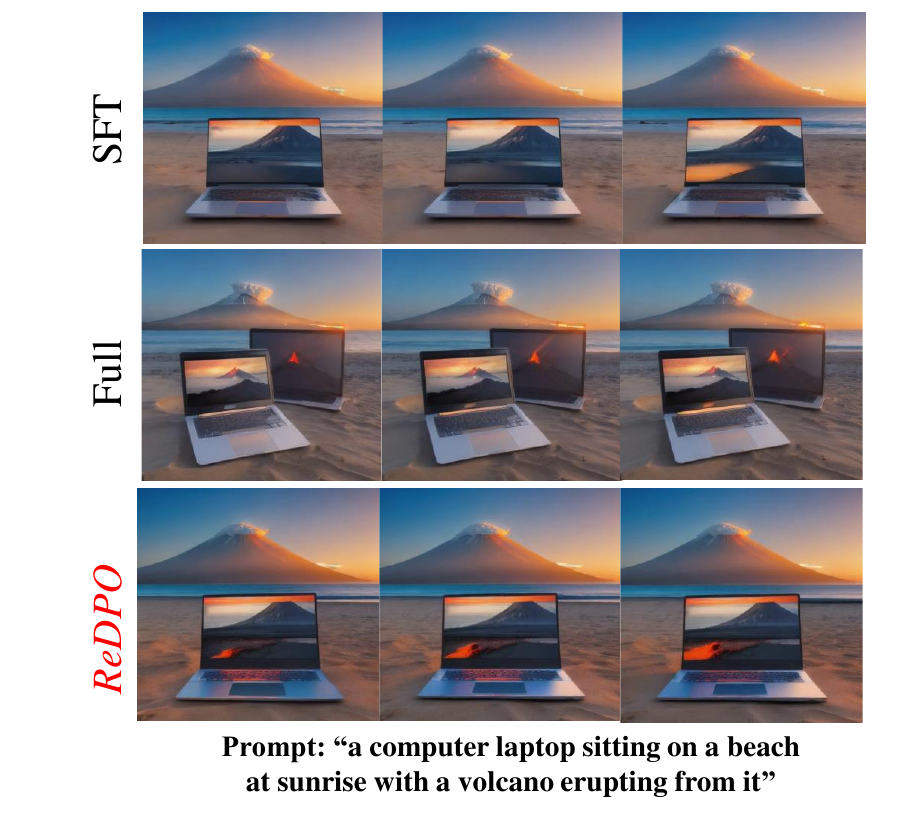}
    \caption{Qualitative example of AnimateDiff}
    \label{fig:ad5}
\end{figure*}

\begin{figure*}[!h]
    \centering
    \includegraphics[width=0.8\textwidth]{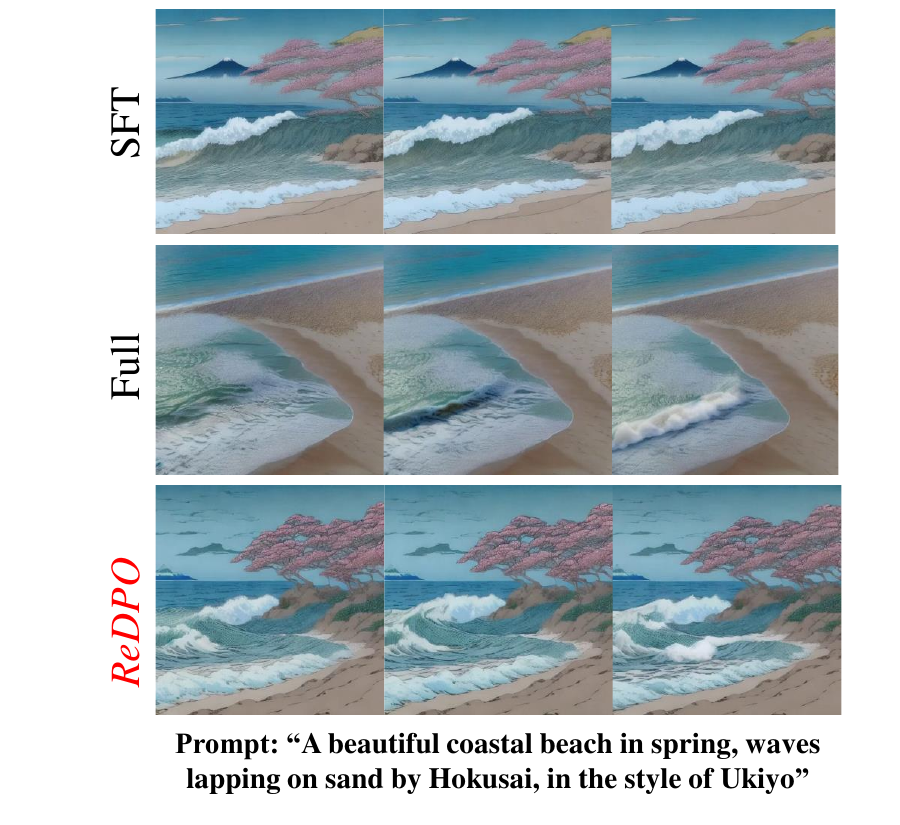}
    \caption{Qualitative example of AnimateDiff}
    \label{fig:ad6}
\end{figure*}

\begin{figure*}[!h]
    \centering
    \includegraphics[width=0.8\textwidth]{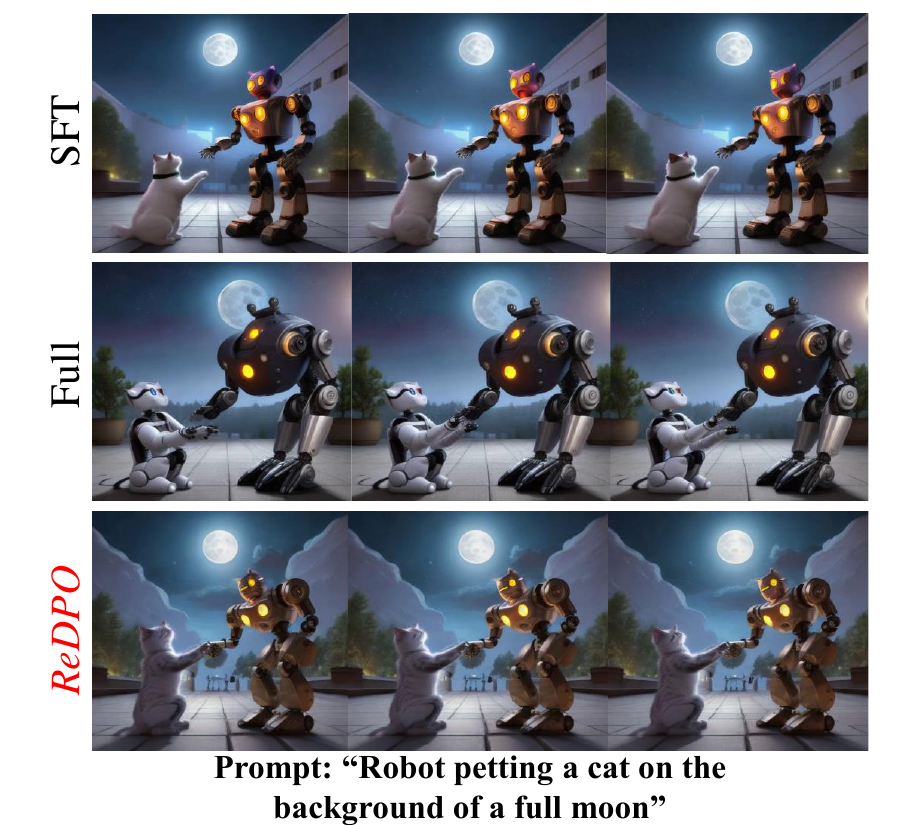}
    \caption{Qualitative example of AnimateDiff}
    \label{fig:ad8}
\end{figure*}

\end{document}